\documentclass[conference]{IEEEtran}

\usepackage{amsmath,amssymb,amsfonts}
\usepackage{algorithm}
\usepackage{algorithmic}
\usepackage{graphicx}
\usepackage{textcomp}
\usepackage{xcolor}
\usepackage{algorithm}
\usepackage{graphicx}
\usepackage{subfigure}
\usepackage{url}
\usepackage{color}
\usepackage{multirow}
\usepackage{xspace}
\usepackage{amsthm}
\usepackage{gensymb}
\usepackage{comment}
\usepackage{xspace}
\usepackage{balance}
\usepackage{lipsum}
\usepackage{cite}

\newcommand{\basealgo}{$\mathsf{LCO}$\xspace}
\newcommand{\basealgobf}{$\boldsymbol{\mathsf{LCO}}$\xspace}

\newcommand{\randombf}{$\boldsymbol{\mathsf{Random}}$\xspace}
\newcommand{\random}{$\mathsf{Random}$\xspace}
\newcommand{\greedybfP}{$\boldsymbol{\mathsf{Greedy}}$\xspace}
\newcommand{\greedyP}{$\mathsf{Greedy}$\xspace}

\newcommand{\ouralg}{{LRCO}\xspace}
\newcommand{\ouralgoP}{{LCRO}\xspace}
\newcommand{\ouralgobfP}{$\boldsymbol{\mathsf{LRCO}}$\xspace}

\newcommand{\weakalgoP}{$\mathsf{Weak\;Oracle}$\xspace}
\newcommand{\weakalgobfP }{$\boldsymbol{\mathsf{Weak\;Oracle}}$\xspace}

\newcommand{\weak}{$\mathsf{Weak\;Oracle}$\xspace}
\newcommand{\oracle}{${\mathsf{Oracle}}$\xspace}

\newcommand{\oraclealgoP}{$\mathsf{Oracle}$\xspace}
\newcommand{\oraclealgobfP }{$\boldsymbol{\mathsf{Oracle}}$\xspace}
\newcommand{\note}[1]{{{#1}}}

\def\BibTeX{{\rm B\kern-.05em{\sc i\kern-.025em b}\kern-.08em
    T\kern-.1667em\lower.7ex\hbox{E}\kern-.125emX}}
\begin{document}

\title{Learning for Robust Combinatorial Optimization: Algorithm and Application
}

\author{
\IEEEauthorblockN{Zhihui Shao}
\IEEEauthorblockA{\textit{UC Riverside}}
\and
\IEEEauthorblockN{Jianyi Yang}
\IEEEauthorblockA{\textit{UC Riverside}}
\and
\IEEEauthorblockN{Cong Shen}
\IEEEauthorblockA{\textit{University of Virginia}}
\and
\IEEEauthorblockN{Shaolei Ren}
\IEEEauthorblockA{\textit{UC Riverside}}
}

\maketitle

\begin{abstract}
Learning to optimize (L2O) has recently emerged as a promising approach to solving
optimization problems by exploiting the strong prediction power of neural networks
and offering lower runtime complexity than conventional
solvers.
While L2O has been applied to various problems, a crucial yet challenging class of problems --- robust
combinatorial optimization in the form of minimax optimization --- have
largely remained under-explored. In addition to the exponentially large
decision space, a key challenge for robust combinatorial optimization
lies in the inner optimization problem, which is typically non-convex
and entangled with outer optimization.
In this paper, we study robust combinatorial optimization and propose
a  novel learning-based optimizer, called \ouralg (Learning for Robust Combinatorial Optimization), which quickly outputs a robust solution in the presence of uncertain context. \ouralg leverages a pair of learning-based optimizers --- one for the minimizer and
the other for the maximizer --- that
use their respective
objective functions as losses and can be trained without the need of labels for training problem instances.
To evaluate the performance of \ouralg,
we perform simulations for
the  task offloading problem in vehicular edge computing.
Our results highlight that \ouralg can greatly reduce the worst-case cost and improve robustness, while having a very low runtime complexity.
\end{abstract}

\section{Introduction}

Fast optimization with competent performance is a fundamental problem in many scientific and engineering fields. Given a problem instance, a typical optimizer
often goes through multiple iterates of time-consuming gradient calculation, thus
having a high computational complexity especially when the problem scales up
\cite{BoydVandenberghe}.
In recent years,
by leveraging the power of neural networks, learning-based optimizers (a.k.a., \emph{learning to optimize},
or simply L2O) have quickly surfaced as efficient and effective solutions to
many optimization problems, complementing or even replacing conventional optimizers \cite{L2O_Scale_Generalize_ICML_2017,L2O_Wireless_PowerAllocation_TSP_2018_8444648,LearningOptimize_PowerAllocation_CongShen_TCOM_2020_8922744,L2O_AdversarialTraining_UCLA_ECCV_2020_l2o_adverarial_train_eccv_2020,L2O_Survey_Benchmark_ZhangyangWang_WotaoYin_arXiv_2021_chen2021learning,L2O_LearningToOptimize_Berkeley_ICLR_2017,L2O_WithGradient_Google_NIPS_2016_andrychowicz2016learning}.
The key idea of L2O is to train a neural network
to learn the optimization process over a set of training problem
instances and generalize to new testing problems.
Compared to conventional optimizers, L2O typically has a lower computational complexity given
a problem instance at runtime  ---  L2O only needs one forward propagation over the learnt
neural network. Importantly,
while ``labels''
(i.e., solutions obtained
by an existing optimizer to training problem instances)
are useful, the training process of L2O can be also directly supervised by
the objective function, whose gradient guides the neural network's parameter updates.
As a result, the training set does not necessarily require labels
\cite{L2O_Survey_Benchmark_ZhangyangWang_WotaoYin_arXiv_2021_chen2021learning,L2O_Combinatorial_Optimization_Survey_Yoshua_2021_BENGIO2021405,L2O_Wireless_PowerAllocation_TSP_2018_8444648,LearningOptimize_PowerAllocation_CongShen_TCOM_2020_8922744}.

Among the family of optimization problems,
combinatorial optimization is a longstanding yet challenging problem with numerous applications, and has received significant attention over the last few decades \cite{Optimization_CombinatorialOptimization_Book_papadimitriou1998combinatorial}. Notable examples range from the classic traveling salesman problem (TSP) and bin-packing problem to
emerging applications such as AI model assignment and server provisioning in edge computing \cite{Edge_Survey_Weisong_Shi_IoTJournal_2016_7488250,L2O_Combinatorial_Survey_Networking_ieee_2020}.
In general, combinatorial optimization is NP-hard due to its exponentially
large solution space and hence often relies on search-based methods (e.g., evolutionary algorithms
\cite{GeneticAlgorithm_Elitist_Model_1996_bhandari1996genetic}),
approximation techniques (e.g., branch
and bound \cite{BalakrishnanBoydBalemi_RobustControl1991}), and/or
heuristics (e.g., greedy-based algorithms).

Recently, L2O has also been explored for combinatorial optimization,
providing competent solutions with a low
computation cost at runtime
\cite{L2O_NeuralCombinatorial_Google_ICLR_Workshop_2017_RL_for_CO_reinforcement_google_2016,L2O_Combinatorial_Optimization_Survey_Yoshua_2021_BENGIO2021405,L2O_Combinatorial_Survey_Networking_ieee_2020,L2O_Combinatorial_Reinforcement_AAAI_2020}.
For example, several studies have proposed L2O-based algorithms
for
various combinatorial problems, including
minimum vertex cover, maximum cut, TSP, mixed-integer programming, and
resource allocation
in communication networks \cite{L2O_Branch_MIP_AAAI_2016,L2O_Combinatorial_Graphs_2017_co_graphs_nips_2017,L2O_Combinatorial_Survey_Networking_ieee_2020,L2O_Scheduling_DRL_Infocom_2019_8737649,L2O_PersonalEdge_ToN_2020_9047133,L2O_Scheduling_Sigcomm_2019_mao2019learning}.

Despite the recent success of L2O,
a crucial class of combinatorial problems --- (worst-case) robust combinatorial optimization
in the form of ``minimax'' ---
have largely remained unexplored. In many practical scenarios, the input/context to the combinatorial
optimization problem is uncertain or even adversarially corrupted.
For example, in the wireless channel assignment problem, the channel
condition for different users is a critical context  but
 can only be estimated subject to estimation errors;
in the cloud resource management problem, the user demand is the context for the resource
 manager, but it can only be predicted
with unavoidable prediction errors;
in the unit commitment problem for power plant scheduling in smart grids,
the electricity demand is provided through a forecast, and hence it cannot be perfectly known a priori and can even be subject to adversarial modification \cite{Baosen_Adversarial_LoadForecasting_SmartGrid_eEnergy_2019_10.1145/3307772.3328314}.
In these applications, simply
viewing the given context as ground truth without accounting for its imperfectness
can result in bad decisions and even catastrophic consequences (e.g., power
outage given an under-estimated electricity demand).
Therefore, it is critically important to take into account the context uncertainty
and achieve robustness in combinatorial optimization.

Nonetheless, robust combinatorial optimization is very challenging, and more so than
the already-difficult combinatorial optimization alone \cite{Optimization_CombinatorialOptimization_Book_papadimitriou1998combinatorial}.
 In fact,
even by relaxing the combinatorial variables
and only considering continuous optimization, robust optimization through
minimax is challenging, unless some additional restrictive assumptions
(e.g., ``convex-concave'' objective function) are imposed \cite{L2O_Robust_L2O_ZhangyangWang_ICLR_2021_shen2021learning}.
A key reason for the challenge is the difficulty in determining the worst case (i.e., the ``$\max$''
part in minimax) that is typically non-convex.
Without a reasonably good solution to the ``$\max$'' part,
the resulting solution may not sufficiently account for the entire uncertainty range
of the context, thus failing to achieve the desired level of robustness.
Moreover, the
maximizer in minimax is also intrinsically entangled with the minimizer.
Thus, standard L2O techniques that assume a single objective function
cannot solve robust combinatorial optimization, where two competing objectives
are entangled.

\textbf{Contribution.} In this paper, we focus on robust minimax combinatorial optimization 
and propose a
novel learning-based optimizer, called \ouralg (Learning for Robust Combinatorial Optimization), which quickly outputs a robust solution in the presence of uncertain context.
As illustrated in Fig.~\ref{fig:system_ren}, \ouralg leverages a pair of learning-based optimizers --- one for the minimizer and
the other for the maximizer --- that interact with each other during the training process and jointly find robust solutions against context uncertainty.
Both the minimizer and maximizer networks use their respective
objective functions as losses which, without the need of labels for training samples,
can directly guide the training process.
Additionally,  we include an ensemble of maximizer networks
to improve the performance of the maximizer.

To evaluate the performance of \ouralg,
we perform numerical experiments for
the  task offloading problem in vehicular edge computing systems. We compare \ouralg with several baseline algorithms and oracle solutions. Our results highlight that \ouralg can greatly reduce the worst-case utility, while also maintaining a good true utility (See Section~\ref{sec:metrics}).
Importantly, \ouralg also has a very low runtime complexity.

\section{Related Works}\label{sec:related_work}
\textbf{Learning to optimize (L2O).} L2O is a general framework
that mimics the optimization process  by training on a set of sample problem instances.
In many of the prior studies, L2O trains a recurrent neural network (RNN)
to parameterize the iteration process (e.g., gradient-based updates)
 \cite{L2O_LearningToOptimize_Berkeley_ICLR_2017,L2O_WithGradient_Google_NIPS_2016_andrychowicz2016learning,L2O_StrongerBaselines_Zhangyang_NIPS_2020_NEURIPS2020_51f4efbf,L2O_Scale_Generalize_ICML_2017,L2O_Survey_Benchmark_ZhangyangWang_WotaoYin_arXiv_2021_chen2021learning}.
For example, concurrent studies \cite{L2O_LearningToOptimize_Berkeley_ICLR_2017,L2O_WithGradient_Google_NIPS_2016_andrychowicz2016learning}
employ RNN-based reinforcement learning (with known objective functions)
to learn the gradient update iterations, while  \cite{L2O_LearnWithoutGraident_ICML_2017_l2o_gradeint_descent_chen_ICML}
subsequently extends the setting to unknown objective functions.
In \cite{L2O_Scale_Generalize_ICML_2017}, a modified RNN is introduced
to improve scalability and generalizability of L2O, and
\cite{L2O_Swarms_ZhangyangWang_NIPS_2019_ensemble_cao_nips_2019}
uses LSTM models and the attention mechanism for solving Bayesian swarm optimization.
To better safeguard neural networks, \cite{L2O_DefenseByLearningToAttack_AISTATS_2021_learning_to_defense_jiang_2018,L2O_AdversarialTraining_UCLA_ECCV_2020_l2o_adverarial_train_eccv_2020}
improve the adversarial training sample generation by using L2O
to solve a constrained maximization problem. In the context of wireless
networks, \cite{L2O_Wireless_PowerAllocation_TSP_2018_8444648,LearningOptimize_PowerAllocation_CongShen_TCOM_2020_8922744} apply L2O to optimize power allocation in multi-user interference channels
for sum rate maximization.
These studies focus on continuous optimization problems and neglect
the uncertainties in the input.

\textbf{Combinatorial optimization.} Traditionally, combinatorial optimization,
such as TSP and bin-packing,
relies on problem-specific heuristics/algorithms designed by domain experts \cite{Combinatorial_book_Korte_2012,Optimization_CombinatorialOptimization_Book_papadimitriou1998combinatorial}.
More recently,
L2O-based algorithms have been proposed
for
a variety of combinatorial problems.
 In general, learning-based combinatorial optimizers can
use label-based supervised learning or label-free reinforcement learning \cite{L2O_Combinatorial_Optimization_Survey_Yoshua_2021_BENGIO2021405}.
In supervised learning, a large number of solutions generated based on state-of-the-art
(SOTA) methods
are used as labels, and the goal of the learnt optimizer is to
directly mimic the solution labels or optimization steps (i.e., imitation learning)
and hence reduce the computational complexity at runtime.
For example,  \cite{PointerNetworks_Combinatorial_Google_NIPS2015_29921001}
proposes pointer networks for combinatorial optimization which are trained
with labels from a SOTA method, and \cite{L2O_Branch_MIP_AAAI_2016}
uses L2O to replace the time-consuming branch and bound method for mixed-integer
programming. By contrast,
methods based on reinforcement learning directly learn the solution
based on the reward/objective function.
 For example, \cite{L2O_NeuralCombinatorial_Google_ICLR_Workshop_2017_RL_for_CO_reinforcement_google_2016}
solves the TSP problem with a close-to-optimal performance, which
is further improved by introducing policy gradient and attention mechanisms
\cite{L2O_TSP_Policy_Gradient_2018_learn_tsp_policy_gradient_2018}.
Learning-based combinatorial optimization  over graphs (e.g., maximum cut
and minimum vertex cover) is studied in
\cite{L2O_Combinatorial_Graphs_2017_co_graphs_nips_2017},
while \cite{L2O_Combinatorial_Survey_Networking_ieee_2020,L2O_Scheduling_DRL_Infocom_2019_8737649,L2O_PersonalEdge_ToN_2020_9047133,L2O_Scheduling_Sigcomm_2019_mao2019learning}
focus on reinforcement learning-based combinatorial optimization problems in networking.
In these studies, the input to the problem instance is assumed to be perfect, and uncertainty of the input is not considered.

\textbf{Minimax optimization.} Robust optimization is very challenging and often formulated
as a minimax problem \cite{Robust_SubmodularMaximization_ICML_2017_bogunovic2017robust,L2O_Robust_L2O_ZhangyangWang_ICLR_2021_shen2021learning}. The conventional solutions to continuous minimax problems are mostly
based on gradient updates or problem-specific methods. For example,
\cite{Minimax_GDA_gradient_minimax_2009} proposes a
gradient descent and gradient ascent (GDA) method, by alternatively
optimizing the $\min$ function and the $\max$ function.
Further, more stable GDA algorithms are also proposed, such as K-Beam \cite{Minimax_KBeam_DeepAdversaial_ICML_2018_pmlr-v80-hamm18a}, optimistic-GDA \cite{Minimax_OGDA_GAN_2017,Minimax_OGDA_Ryu_arXiv_2019} and Follow-the-Ridge \cite{Minimax_FollowRidge_ICLR_2020_Wang2020On}.

A generative adversarial network also involves minimax optimization
\cite{DNN_GAN_Goodfellow_NIPS_2014_5ca3e9b1}, but it typically
requires true samples/labels to train
the discriminator for distinguishing fake samples from true ones.
By contrast, \ouralg learns two optimizers ($\min$ and $\max$),
and the notion of ``true'' samples does not apply.

Because of its combinatorial nature, robust combinatorial optimization has remained relatively under-explored, with some non-learning-based algorithms available under specific settings
and assumptions (e.g., \cite{Robust_MonotoneSubmodular_Math_2018_orlin2018robust,Robust_SubmodularMaximization_ICML_2017_bogunovic2017robust}
consider monotone and submodular functions with a robust setting where
a certain number of actions can be nullified).

The very recent study \cite{L2O_Robust_L2O_ZhangyangWang_ICLR_2021_shen2021learning}
uses L2O to solve \emph{continuous} minimax problems, but our work is substantially different.
Concretely, \cite{L2O_Robust_L2O_ZhangyangWang_ICLR_2021_shen2021learning}
trains two individual LSTM networks  using a self-defined loss function,
and uses these two networks to replace the gradient update process for both the $\min$ and $\max$ parts (used in conventional optimizers like \cite{Minimax_GDA_gradient_minimax_2009}).
Thus, this approach does not apply to robust combinatorial optimization which,
unlike continuous optimization,
does not have standard gradient updates and is more challenging. Additionally,
unlike \cite{L2O_Robust_L2O_ZhangyangWang_ICLR_2021_shen2021learning}
using a single network for the $\max$ part, we include
an ensemble of networks to further improve the maximizer performance.

In summary, \ouralg advances the quickly expanding field of L2O
and is the
first learning-based optimizer offering a fast and better solution to \emph{robust} combinatorial optimization.

\section{Problem Formulation}\label{sec:problem}
Combinatorial optimization problems widely exist in scientific and engineering applications, including resource allocation, scheduling, and capacity planning in computing and networking systems \cite{L2O_Combinatorial_Survey_Networking_ieee_2020}.
In general, a combinatorial optimization problem can be written as
$\underset{\mathbf{a}\in\mathcal{A}}{\text{min}} \; f(\mathbf{x},\mathbf{a})$,
where $\mathbf{a}$ is the decision variable, $\mathcal{A}$ is the feasible decision set, and $f(\cdot,\cdot)$ indicates the objective function with the context parameter $\mathbf{x}$
provided to the optimizer.
For generality, we note that both the context $\mathbf{x}$ and decision $\mathbf{a}$ are \emph{vectors} but  with possibly different dimensions.
This formulation applies to integer programming, which also belongs to combinatorial optimization.

A key novelty that distinguishes our work apart from standard combinatorial optimization
is consideration of the uncertainty in the context parameter $\mathbf{x}$, as motivated by practical applications. For example, before deciding the optimal channel assignment for users
in a wireless network, the channel condition (i.e., context $\mathbf{x}$ in our formulation)
is needed, but it can only be estimated. To capture the context uncertainty,
we assume that the true context is $\mathbf{x}+\delta$, where the context error $\delta$
belongs to an uncertainty set $\Delta$ (also referred to as uncertainty/error budget in robust optimization). In the robust optimization and learning literature \cite{L2O_Robust_L2O_ZhangyangWang_ICLR_2021_shen2021learning,Adversarial_AdversarialML_Berkeley_AISec_2011_10.1145/2046684.2046692}, $\delta$ is commonly bounded by a $L_p$ norm,
i.e., $\Delta=\{\delta,\;|\delta|_p\leq\epsilon\}$ where $|\delta|_p$ denotes
the $L_p$ norm of $\delta$ with $p\geq1$.
While we consider continuous context parameters
$\mathbf{x}$ for the ease of presentation,
\ouralg can also be extended to discrete contexts by modifying the learning-based maximizer
(Section~\ref{sec:algorithm_maximizer_critic}) such that it solves combinatorial optimization.

We now formulate the robust combinatorial optimization problem as follows:
\begin{equation}\label{eqn:opt_robust}
\underset{\mathbf{a}\in\mathcal{A}}{\min} \; \underset{\delta\in\Delta}{\max} \; f(\mathbf{x}+\delta,\mathbf{a})
\end{equation}
where $\mathbf{x}$ is the uncertain context available to the optimizer and $\delta\in\Delta$ denotes the context error unknown to the optimizer.
Our minimax formulation captures the \emph{worst-case} robustness over the entire uncertainty set $\Delta$, and is commonly used as a robust metric \cite{L2O_Robust_L2O_ZhangyangWang_ICLR_2021_shen2021learning,Robust_MonotoneSubmodular_Math_2018_orlin2018robust,Robust_SubmodularMaximization_ICML_2017_bogunovic2017robust}.
Note that the actually achieved cost may not be the worst case and hence can often be lower
than worst-case cost.

\textbf{Challenges.}
Even with the perfect context (i.e., $\delta=0$),
combinatorial optimization is in general NP-hard and can only be solved by approximate algorithms \cite{Optimization_CombinatorialOptimization_Book_papadimitriou1998combinatorial,Combinatorial_book_Korte_2012,L2O_Combinatorial_Survey_Networking_ieee_2020}.
Furthermore, the uncertain context $\delta\in\Delta$
makes robust combinatorial optimization even more challenging, because it is time-consuming to use a conventional solver to solve the inner maximization problem in Eqn.~\eqref{eqn:opt_robust} due to its non-convexity in many practical applications.
Thus, except for special cases, it is generally impossible to have a closed-form solution for the inner problem $\underset{\delta\in\Delta}{\max} \; f(\mathbf{x}+\delta,\mathbf{a})$.
As a result,
the outer minimization problem cannot be efficiently solved either.
Even though we can use a learning-based minimizer \cite{L2O_Combinatorial_Graphs_2017_co_graphs_nips_2017,L2O_Combinatorial_Survey_Networking_ieee_2020} along with a conventional
maximizer, the runtime computational complexity given a new problem instance
is very high, because the learning-based minimizer produces a probability distribution
for decisions and the conventional maximizer needs to be executed
once for each sampled decision
in order to select the optimal robust decision.
As a preview of the numerical results (Section~\ref{sec:vcc_exmaple}), the solver is about 70x slower than our learning-based
maximizer, and the maximizer needs to find the worst-case cost for each of the 1000 candidate decisions sampled by the minimizer.
On the other hand, while simple heuristic methods (e.g., greedy) can be fast, they may not offer a satisfactory
performance.
To address these challenges,
we shall present \ouralg, a learning-based optimizer that produces a competent solution $\mathbf{a}$ with a low computational complexity at runtime.
\begin{figure}
	\centering
	\includegraphics[trim=0cm 12cm 31.5cm 0cm,clip,  width=0.47\textwidth]{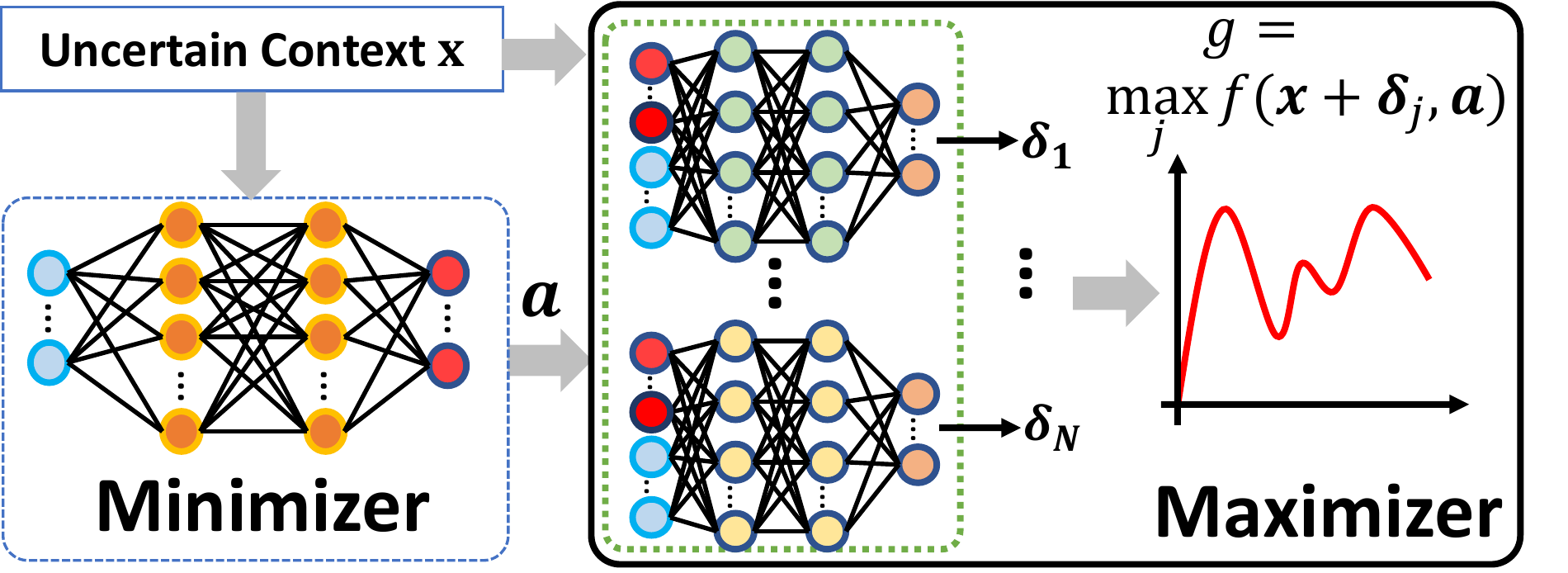}
	\caption{Overview of \ouralg. The maximizer network produces
a set of candidate actions, which are then evaluated by
the minimizer for robustness. The action with the minimum
worst-case cost is selected.}\label{fig:system_ren}
\end{figure}

\section{\ouralg: Learning for Robust Combinatorial Optimization}\label{sec:method}

In this section, we provide the design of \ouralg, which combines two learning-based optimizers (i.e., $\min$ and $\max$) for efficiently solving robust optimization problems.

\subsection{Overview}

The key idea of \ouralg is to use machine learning models,
in particular neural networks, to parameterize
the minimizer and maximizer for improving computational efficiency given new problem instances
at the runtime. We choose neural networks mainly because of their strong universal
approximation capability \cite{DNN_Universal_Apprixmator_1989_10.5555/70405.70408}.
An overview of \ouralg is shown in Fig.~\ref{fig:system_ren}, which includes
two learning-based optimizers --- a minimizer and a maximizer that solve
the outer and inner part of Eqn.~\eqref{eqn:opt_robust}, respectively.

\textbf{Testing/inference.} Given a problem instance with uncertain context
at runtime, the  optimization process is described in Algorithm~\ref{alg:testing}.
First,
the neural network-based minimizer
outputs the probability of decisions, based on which a number of candidate decisions are sampled accordingly.
For the maximizer which typically involves non-convex optimization,
we use an ensemble of neural networks.
Specifically, given each pair of uncertain
context $\mathbf{x}$ and sampled decision $\mathbf{a}$ (produced by the minimizer),
the maximizer obtains  the
the worst-case context error $\delta$ through the ensemble.
Finally, the decision that has the minimum worst-case cost is chosen.
At runtime, the entire optimization process in \ouralg only involves forward propagation over
the minimizer and maximizer neural networks, and hence has a much lower complexity
than using conventional solvers.

\begin{algorithm}[!t]
	\caption{\ouralg for Testing/Inference}\label{alg:testing}
	\begin{algorithmic}[1]
		\STATE $\textbf{Inputs:}$ Uncertain context $\mathbf{x}$,
 minimizer network  parameter $\theta_a$, and  ensemble
 maximizer network parameters
$\theta_{w,i}$ for $i=1,\cdots,N$
        \STATE $\textbf{Output:}$ Robust decision $\mathbf{a}$
		\STATE{Use the minimizer network to generate  action probability distribution based on Eqn.~\eqref{eqn:prob_factor}, and sample $K$ candidate decisions $\mathbf{a}_k$ accordingly}
        \FOR {$k=1,\cdots,K$}
        \FOR {$i=1,\cdots,N$}
		\STATE{Use the maximizer
network $i$ to generate the worst-case context error $\delta_{i,k}$
for decision $\mathbf{a}_k$}	
        \STATE{Calculate $g_{i,k}=f(\mathbf{x}+\delta_{i,k},\mathbf{a}_k)$}
		\ENDFOR
        \STATE{Find the worst-case cost $G_k=\max_i g_{i,k}$ for $\mathbf{a}_k$}	
		\ENDFOR	
        \STATE{Select the decision $\mathbf{a}=\mathbf{a}_k$, where
        $k=\arg\min_kG_k$}
	\end{algorithmic}
\end{algorithm}	

\textbf{Training.} For training the neural networks, we do not provide ground-truth sample solutions as labels, i.e., solutions
obtained by an existing solver. One reason is that it can be very time-consuming to generate solutions
to many problem instances (even offline) due to the combinatorial nature in the minimizer
and non-concavity in the maximizer. In addition, for complex optimization problems,
 solutions directly learnt by neural networks may even exceed
 those obtained by
 conventional solvers \cite{LearningOptimize_PowerAllocation_CongShen_TCOM_2020_8922744}.
Thus, we use the objective function $f(\cdot,\cdot)$ as the loss
to directly supervise the training.
Note that, like other machine learning methods, the goal
of \ouralg is to speed up the runtime inference for testing samples (compared to non-learning methods), although it requires additional offline training time.

\textbf{\ouralg vs. reinforcement learning.} The architecture of \ouralg resembles the actor-critic reinforcement learning \cite{Reinforcement_Book_2018)sutton2018reinforcement_intro}: in \ouralg, the minimizer's
decision is evaluated by the maximizer, while the actor's policy is evaluated by the
critic in reinforcement learning. Nonetheless, there are also crucial differences.
First, in \ouralg, the training is performed entirely offline based on the distribution of context $\mathbf{x}$ since the objective function $f(\cdot,\cdot)$ is already known,
whereas reinforcement learning is often updated online to adapt to a new environment.
Second, the actor in reinforcement learning typically learns the value function
based on reward signal feedback (i.e., labels), whereas the maximizer in \ouralg leverages
L2O to solve an (often non-convex) optimization problem.

\subsection{Maximizer Network}\label{sec:algorithm_maximizer_critic}

We now present the details of our learning-based maximizer, which includes
an ensemble of neural networks to efficiently produce a solution to the inner maximization
problem in ${\min}_{\mathbf{a}\in\mathcal{A}}  {\max}_{\delta\in\Delta} \; f(\mathbf{x}+\delta,\mathbf{a})$.

In general, the inner maximization problem is non-convex (e.g., the vehicular task
offloading problem in Section~\ref{sec:vcc_exmaple}). Alternatively,
one may want to use a conventional solver (e.g., based on gradient updates \cite{BoydVandenberghe}) for the inner maximization problem,
but this has a high computational cost.
Specifically, for testing at runtime,
the minimizer produces a probability distribution
for candidate decisions, and the maximizer needs
to evaluate the worst-case cost for each sampled decision (e.g., 1000 in total
in Section~\ref{sec:vcc_exmaple}) in order to select the optimal decision.
Thus, using conventional solvers can incur a very high computational complexity.

In \ouralg, we  train the maximizer network by directly solving the inner optimization problem $\underset{\delta\in\Delta}{\max} \; f(\mathbf{x}+\delta,\mathbf{a})$ where $\Delta=\{\delta,\;|\delta|_p\leq\epsilon\}$ is the uncertainty set.
Specifically, by viewing the uncertain context $\mathbf{x}$ and (candidate) decision $\mathbf{a}$
as the input, the maximizer network learns to optimize $\delta$, as supervised
by the training loss function below:
\begin{equation}\label{eqn:inner_max_loss}
\mathcal{L}(\mathbf{x},\mathbf{a},\delta) = -f(\mathbf{x}+\delta,\mathbf{a}) + \lambda \left[|\delta|_p - \epsilon\right]^+,
\end{equation}
where we add the minus sign to be consistent with the definition of loss function,
and $\lambda\left[|\delta|_p - \epsilon\right]^+$ means that additional penalty weighted by
$\lambda>0$ is imposed
whenever the solution $\delta$ goes beyond the uncertainty set $\Delta$.
As the neural network is trained based on sampled problem instances,
 the maximizer network can ensure
by tuning $\lambda$
that the $\left[|\delta|_p - \epsilon\right]^+$ term is sufficiently small on \emph{average}.
In case that the output $\delta$ violates the uncertainty set $\Delta$ for a particular testing sample,
we can scale down $\delta$ to force it to fall into $\Delta$.

\begin{algorithm}[!t]
	\caption{Iterative training of \ouralg}\label{alg:iterative_training}
	\begin{algorithmic}[1]
		\STATE $\textbf{Inputs:}$ Training set of uncertain context $\mathcal{D}_x$,
        \STATE {$\textbf{Output:}$ Minimizer network  parameter $\theta_a$, and  ensemble
 maximizer network parameters
$\theta_{w,i}$ for $i=1,\cdots,N$}
		\STATE{$\textbf{Initialize:}$ Randomly generate a training set of decisions $\mathcal{D}^0_{\mathbf{a}}$, and
 pre-train the ensemble of maximizer networks $\theta_{w,i}$ for $i=1,\cdots,N$}
 over $\mathcal{D}_x$ and $\mathcal{D}^0_{\mathbf{a}}$
        \FOR {$k=1,\cdots,MaxIterate$}
        \STATE Train the minimizer network using Algorithm~\ref{alg:minimizer_training}
        \FOR {$\mathbf{x}\in \mathcal{D}_x$}
        \STATE Calculate $P_{\theta_a}(\textbf{a}|\mathbf{x})$ based on Eqn.~\eqref{eqn:prob_factor}
		\ENDFOR
        \STATE Calculate the new distribution of decisions $P_{\theta_a}(\textbf{a})=\frac{1}{|\mathcal{D}_x|}\sum{x\in \mathcal{D}_x}P_{\theta_a}(\textbf{a}|\mathbf{x})$	
        \STATE Randomly generate a new training set of decisions $\mathcal{D}^k_{\mathbf{a}}$ based on the new distribution $P_{\theta_a}(\textbf{a})$
        \STATE Train the ensemble of maximizer networks $\theta_{w,i}$, for $i=1,\cdots,N$, over the dataset $\mathcal{D}_x$ and $\mathcal{D}^k_{\mathbf{a}}$
        \ENDFOR
	\end{algorithmic}
\end{algorithm}

To train the maximizer network, we first generate samples based on the distribution
of uncertain context $\mathbf{x}$ and candidate decisions $\mathbf{a}$. Then, with training samples and the loss function in Eqn~\eqref{eqn:inner_max_loss}, any standard learning approaches, like stochastic gradient descent, can be applied, and hence we omit the description.
Since  the inner optimization problem $\underset{\delta\in\Delta}{\max}  f(\mathbf{x}+\delta,\mathbf{a})$ is typically non-convex, simply using one maximizer network
may not produce a satisfactory solution. Thus, for the maximizer, we use an ensemble of neural networks, each initialized with a different weight.
Prior studies \cite{LearningOptimize_PowerAllocation_CongShen_TCOM_2020_8922744} have
shown that the ensemble approach can significantly improve the optimization performance.

We note that the distribution
of decisions $\mathbf{a}$ is entangled with the minimizer network's output. One approach
is to randomly generate candidate decisions covering a sufficiently wide distribution. While the resulting maximizer network
can produce good solutions to a wide range of decision distributions,
the actual distribution of the decisions produced by the minimizer network
is only a subset of the distribution used for training. In other words,
the capacity of the maximizer neural networks is not fully utilized.
Here, we first randomly generate  $\mathbf{a}$ and pre-train the minimizer network, use it to evaluate the worst-case to guide
the minimizer network's training, and then update the training set of
$\mathbf{a}$ based on the new distribution of decisions produced by the minimizer network.
This process can repeat a few iterations as shown in Algorithm~\ref{alg:iterative_training}.
\note{While we observe convergence in all our empirical experiments,
 it is an interesting future study to derive rigorous
conditions under which the iterative training
process is guaranteed to converge.}

\subsection{Minimizer Network}
The learning-based minimizer employs a
neural network that parameterizes the solution to the outer combinatorial optimization
problem in $\underset{\mathbf{a}\in\mathcal{A}}{\min} \; \underset{\delta\in\Delta}{\max} \; f(\mathbf{x}+\delta,\mathbf{a})$. Instead of generating labels (i.e., solutions obtained by an existing solver)
for supervision, we use a reinforcement learning process to train
the minimizer network. Although there exist
many reinforcement learning-based combinatorial optimizers
\cite{L2O_NeuralCombinatorial_Google_ICLR_Workshop_2017_RL_for_CO_reinforcement_google_2016,L2O_Combinatorial_Optimization_Survey_Yoshua_2021_BENGIO2021405,L2O_Combinatorial_Survey_Networking_ieee_2020,L2O_Combinatorial_Reinforcement_AAAI_2020},
they are directly
supervised by the objective function
under the assumption of perfect context parameters. In sharp contrast,
to account for uncertain context parameters, the minimizer network in \ouralg uses the worst-case cost returned
by our learning-based maximizer (Section~\ref{sec:algorithm_maximizer_critic})
for supervision. Additionally, due to the coupling of minimization and maximization
in ${\min}_{\mathbf{a}\in\mathcal{A}}  {\max}_{\delta\in\Delta} \; f(\mathbf{x}+\delta,\mathbf{a})$,
the output of our minimizer network also influences
the training of the minimizer (see Algorithm~\ref{alg:iterative_training}).

Concretely, the minimizer network parameterized by $\theta_a$ takes the uncertain context
$\mathbf{x}$ as input and yields the probability of candidate decisions
$p_{\theta_a}(\mathbf{a}|x)$ over the feasible decision set
$\mathcal{A}$. The parameter $\theta_a$ is trained by following
a policy gradient algorithm under the supervision of our minimizer
(which plays a similar role of the ``critic'' in reinforcement learning) \cite{Reinforcement_Book_2018)sutton2018reinforcement_intro}.
The training process is described in Algorithm~\ref{alg:minimizer_training}.

\textbf{Decision representation.}
A well-trained minimizer network should assign higher probabilities to better decisions that are more likely to minimize the worst-case cost. For combinatorial optimization,
the feasible decision set is typically exponentially large. For example,
a 10-dimensional decision $\mathbf{a}$, with 5 possible integer values for each dimension, will  result in $5^{10}$ decisions.
Thus, it is difficult to directly represent all the decisions using
the conventional one-hot encoding, where each node in the output layer represents a single possible decision. To address this issue,
we divide the decision $\mathbf{a}$ into $D$ decision groups. 
Within each group $i$, we use an integer $a_i$ to represent the feasible decisions
using one-hot encoding in the final output layer, followed by softmax activation which
produces the probability distribution for each decision group.\footnote{If a decision group only contains one binary decision, we can then
use only one output node followed by sigmoid activation for that decision group.}
Then, by rewriting the decision vector as $\mathbf{a}=[a_1,\cdots,a_D]^T$, we
factorize the final decision probability as follows: 
\begin{equation}\label{eqn:prob_factor}
P_{\theta_a}(\mathbf{a}|x)=\prod_{i=1}^{D}p_{\theta_a}(a_i|\mathbf{x}).
\end{equation}
To further illustrate this point, consider that we have 10 communication channels
and need to decide whether to occupy each channel. For this case, we have
10 independent decision groups, each with a binary decision.
If it is not possible to divide the decision $\mathbf{a}$ into {independent} decision groups,
 we can temporarily ignore the coupling constraint and then re-scale the decision probability. For example, using the previous example
 and assuming a cardinality constraint $B$ on the set of selected channels,
 we can obtain $P_{\theta_a}(\mathbf{a}|\mathbf{x})=\prod_{i=1}^{D}p_{\theta_a}(a_i|\mathbf{x})$ and only
 consider those decisions with the $B$ highest probabilities.
While we present a general approach, we can also exploit applicable problem structures, e.g., graph topology,
to better encode the decisions (see \cite{L2O_Combinatorial_Graphs_2017_co_graphs_nips_2017}
for an example).

\begin{algorithm}[!t]
	\caption{Minimizer Network Training}\label{alg:minimizer_training}
	\begin{algorithmic}[1]
		\STATE $\textbf{Inputs:}$ Training set of uncertain context $\mathcal{D}_x$,
and ensemble network for the maximizer, training epochs $T$, batch size $B$, sampling size $|\mathcal{S}|$
        \STATE {$\textbf{Output:}$ Minimizer network  weight parameter $\theta_a$}
		\STATE{$\textbf{Initialize:}$ Randomly initialize the parameter $\theta_a$}
        \FOR {$t=1,\cdots,T$}
        \FOR {$i= {1,\cdots,B}$}
        \STATE{Randomly sample $\mathbf{x}$ from $\mathcal{D}_x$}
        \STATE{{Calculate decision probability $P_{\theta_a}(\mathbf{a}|\mathbf{x})$ based on Eqn.~\eqref{eqn:prob_factor}, and sample decisions from $P_{\theta_a}(\mathbf{a}|\mathbf{x})$}}
        \STATE{Run the maximizer (Section~\ref{sec:algorithm_maximizer_critic})
        to find the worst-case cost $G=\max_{\delta\in\Delta} f(\mathbf{x}+\delta,\mathbf{a})$}
        \STATE{Generate a random set of decisions $\mathcal{S}$ from $\mathcal{A}$
        }
        \FOR{$\mathbf{d}_s\in \mathcal{S}$}
        \STATE Run the maximizer
        to find the worst-case cost $G_s=\max_{\delta\in\Delta} f(\mathbf{x}+\delta,\mathbf{a}_s)$
        \ENDFOR
        \STATE {Calculate the baseline $V(\mathbf{x})=\frac{1}{|\mathcal{S}|}\sum_{s=1}^{|\mathcal{S}|} G_s$}
        \STATE{{Calculate gradient $-\left[G-V(\mathbf{x})\right]\nabla_{\theta_a} P_{\theta_a}(\mathbf{a}|\mathbf{x})$}}
        \ENDFOR
        \STATE{Update $\theta_a$ with batched gradient}
        \ENDFOR
	\end{algorithmic}
\end{algorithm}	

\textbf{Policy gradient.}
As in the Monte-Carlo reinforced policy gradient approach \cite{Reinforcement_Book_2018)sutton2018reinforcement_intro,L2O_NeuralCombinatorial_Google_ICLR_Workshop_2017_RL_for_CO_reinforcement_google_2016}, the net cost benefit of a decision $\max_{\delta\in\Delta} f(\mathbf{x}+\delta,\mathbf{a})-V(\mathbf{x})$
is used for gradient calculation, instead of the absolute cost $\max_{\delta\in\Delta} f(\mathbf{x}+\delta,\mathbf{a})$. Here, $V(\mathbf{x})$ is the baseline value function calculated based on
a set of randomly sampled decisions  (Line~13 of Algorithm~\ref{alg:minimizer_training}).
This can effectively avoid unnecessarily penalizing those decisions that are good (relative to the baseline) but have low absolute values. The sampling process in Line~7 of
of Algorithm~\ref{alg:minimizer_training} also encourages exploration without always exploitation.

We can also employ an ensemble of neural networks for the minimizer,
although our results in Section~\ref{sec:vcc_exmaple} shows that a single neural network has already performed very well.
\section{Application: Task Offloading in Vehicular Edge Computing}\label{sec:vcc_exmaple}
To evaluate \ouralg in real networking applications, we apply \ouralg to the problem of replicated task offloading  in a vehicular edge computing (VEC) system.
We first describe the problem formulation, then explain the simulation setup, and finally
present the results under two different error budgets.

\subsection{Background and Problem Formulation}
With the advances in  connected driving
and vehicular networks, VEC is emerging as a promising computing architecture, complementing the conventional cloud systems.
Specifically, to improve the service quality and user experience,  computation is gradually moving toward the edge of vehicular networks, and vehicles with extra computation resources can even provide distributed computing to other vehicles and nodes. We refer
to those vehicles that can provide computation on the move
as \emph{vehicular clouds}. Nonetheless,
because of  the continuously changing environment (e.g, vehicle's locations and/or server resource availability), the resource management decisions in a VEC system
must also be highly agile and adaptive
 --- the resource manager needs to dynamically and efficiently assign the workloads to different servers. Furthermore,
 performance robustness is crucial in VEC, especially for safety-critical applications.
Next,  we consider the replicated task offloading problem in VEC and formulate
it into robust combinatorial optimization.

\begin{figure}[!h]
	\centering
	\includegraphics[trim=1cm 9cm 8cm 0cm,clip,  width=0.45\textwidth]{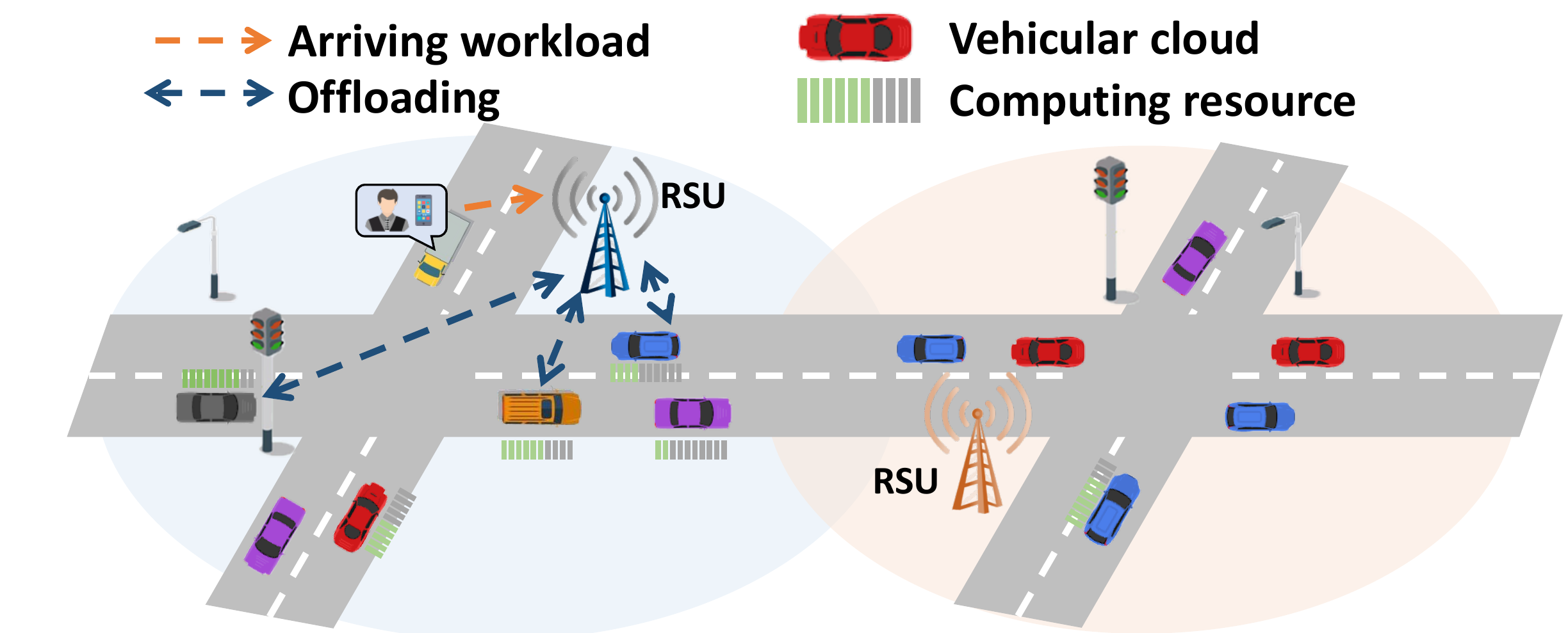}
	\caption{Illustration of task offloading in a VEC system.}\label{fig:VEC_diagram}
\end{figure}

Consider a VEC system as illustrated in Fig.~\ref{fig:VEC_diagram}, including a road-side-unit (RSU) and multiple vehicular clouds that can provide computation on demand. The RSU is installed at a fixed location, which gathers computing demand/tasks from some vehicles
and then offloads them to nearby available vehicular clouds via wireless communications. Each workload includes multiple micro services, which are offloaded to different vehicular clouds. The service delay $d$ for each micro service includes the data transmission delay $d^t$ of the wireless network and service computing delay $d^c$ in the respective vehicular cloud. By default, each task arrives with a maximum delay constraint $L$. The task
is successfully finished if all involved micro services are completed prior to the deadline, and fails otherwise.

Due to the highly dynamic environment (e.g., fast-changing locations of vehicular clouds), it is difficult to guarantee the success  of each offloaded task, and estimating the success rate of each offloading decision a priori is also non-trivial.
Thus,
to achieve a higher success rate, each arriving task is replicated and offloaded to multiple vehicular clouds as in \cite{vcc_seurity_2012_IEEE_c4,xu_jie_2018_infocom_task_replication_c4}.
In addition, it is important to design a robust offloading policy for the RSU, improving the overall success rate even in the worst case. We denote $x_{ij}$ as the success rate when a micro service $j$ is offloaded to vehicular cloud $v_i$. Then, the overall task success rate can be calculated via the \textit{At-Least-One} rule as shown below:
\begin{equation}\label{eqn:overall_prob}
P(\textbf{x},\textbf{a}) = \prod_{j=1}^M\left[1-\prod_{i=1}^{C}(1-x_{ij} a_{ij})\right]
\end{equation}
where $a_{ij}\in\{1,0\}$ represents
the offloading decision  (i.e.,
``1'' means offloading the micro service $j$ to
vehicular cloud $i$, and ``0'' means otherwise),
 $M$ is the total number of micro services in the task,
 and $C$ is the maximum number of vehicular clouds to which each micro service can be offloaded.
The interpretation of Eqn.~\eqref{eqn:overall_prob}
is that each micro service should be successfully completed by at least one vehicular cloud,
and the overall task is successful only when all the micro services are successfully
executed.

\begin{figure*}[!t]
	\centering
	\subfigure[$L=0.25$]{\includegraphics[trim=0cm 0.5cm 0cm 0cm,clip,  width=0.23\textwidth]{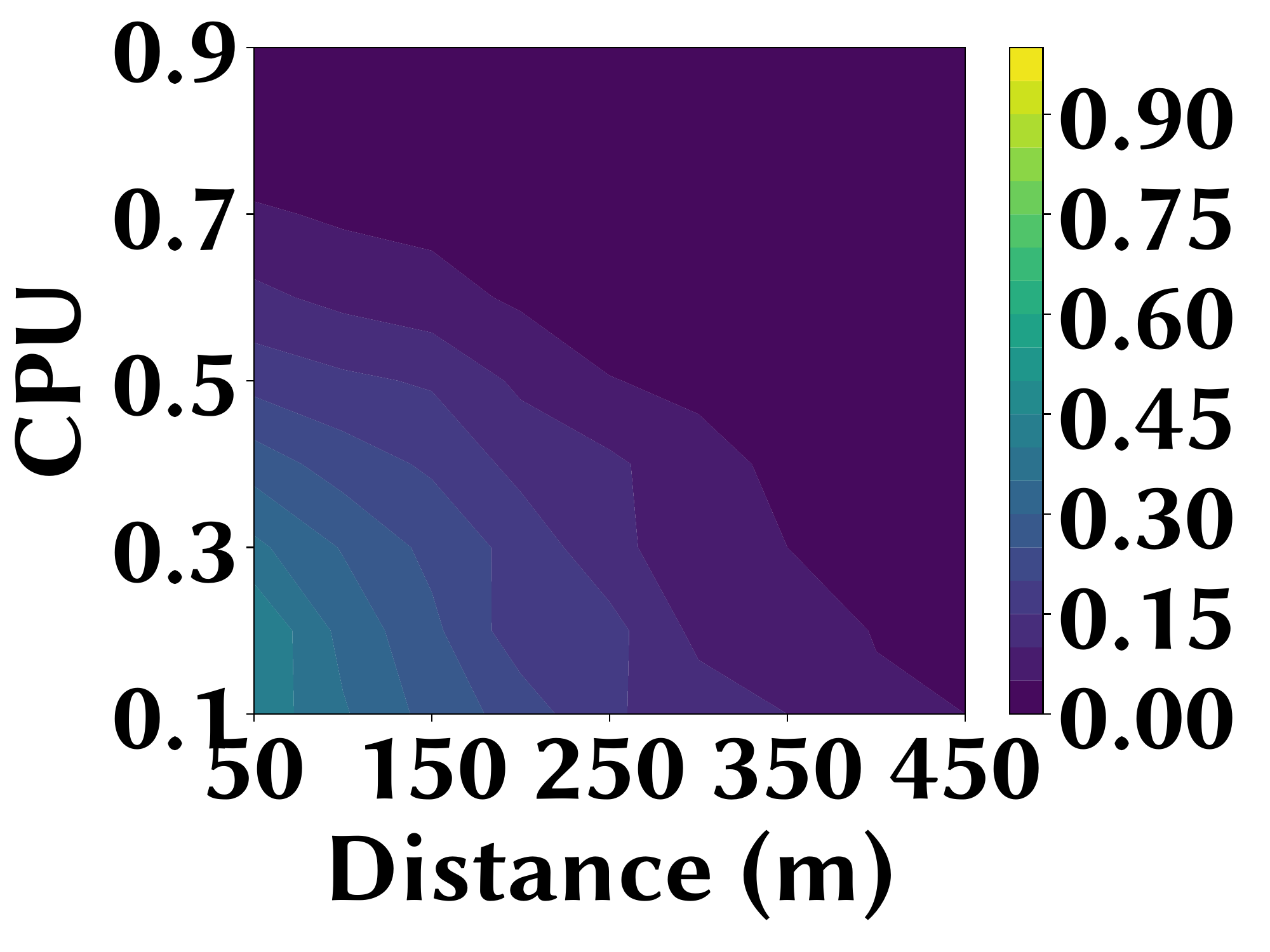}\label{fig:dataset_lt_1}}
	\subfigure[$L=0.5$]{\includegraphics[trim=0cm 0.5cm 0cm 0cm,clip,  width=0.23\textwidth]{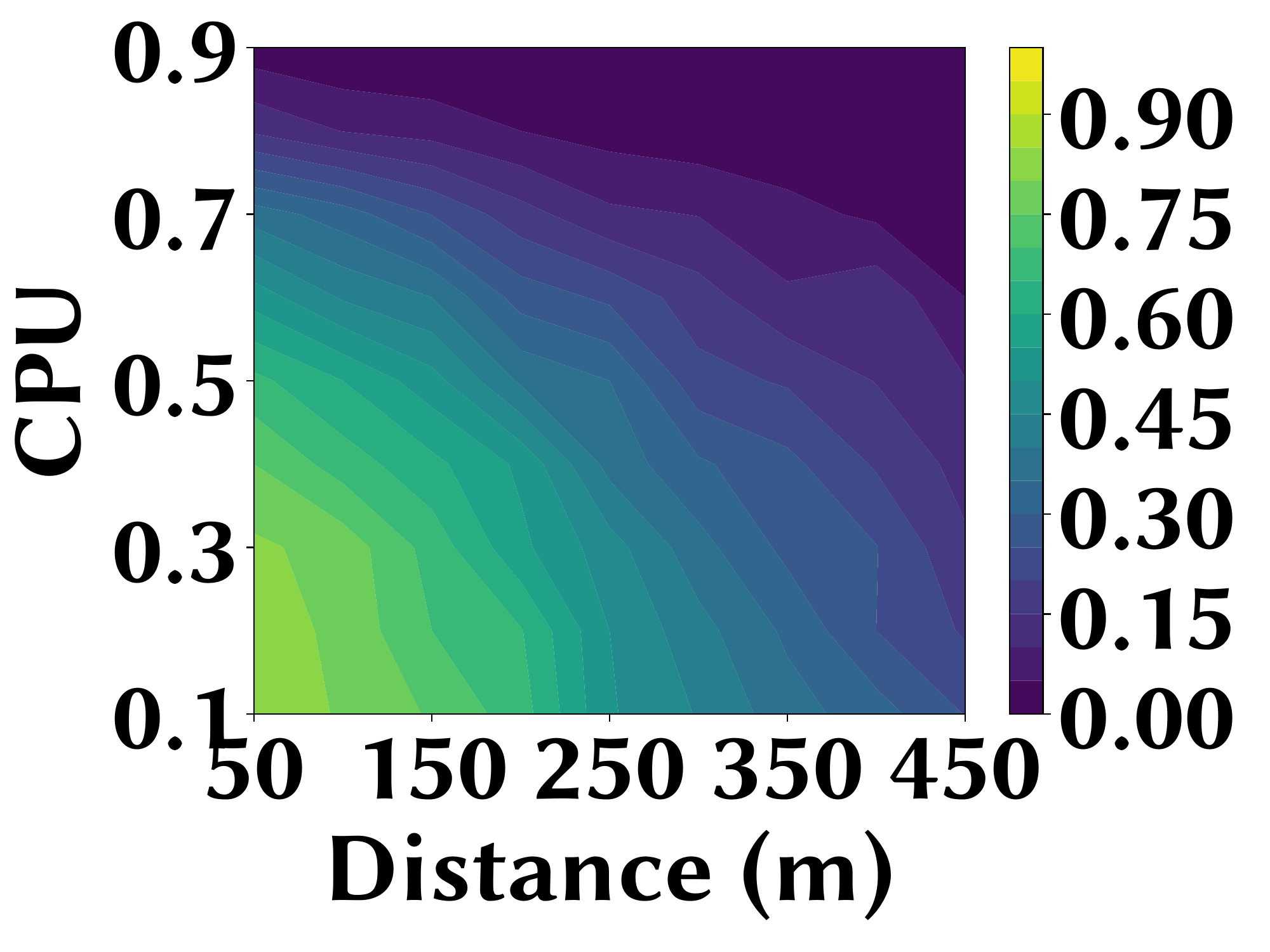}\label{fig:dataset_lt_2}}
	\subfigure[$L=0.75$]{\includegraphics[trim=0cm 0.5cm 0cm 0cm,clip,  width=0.23\textwidth]{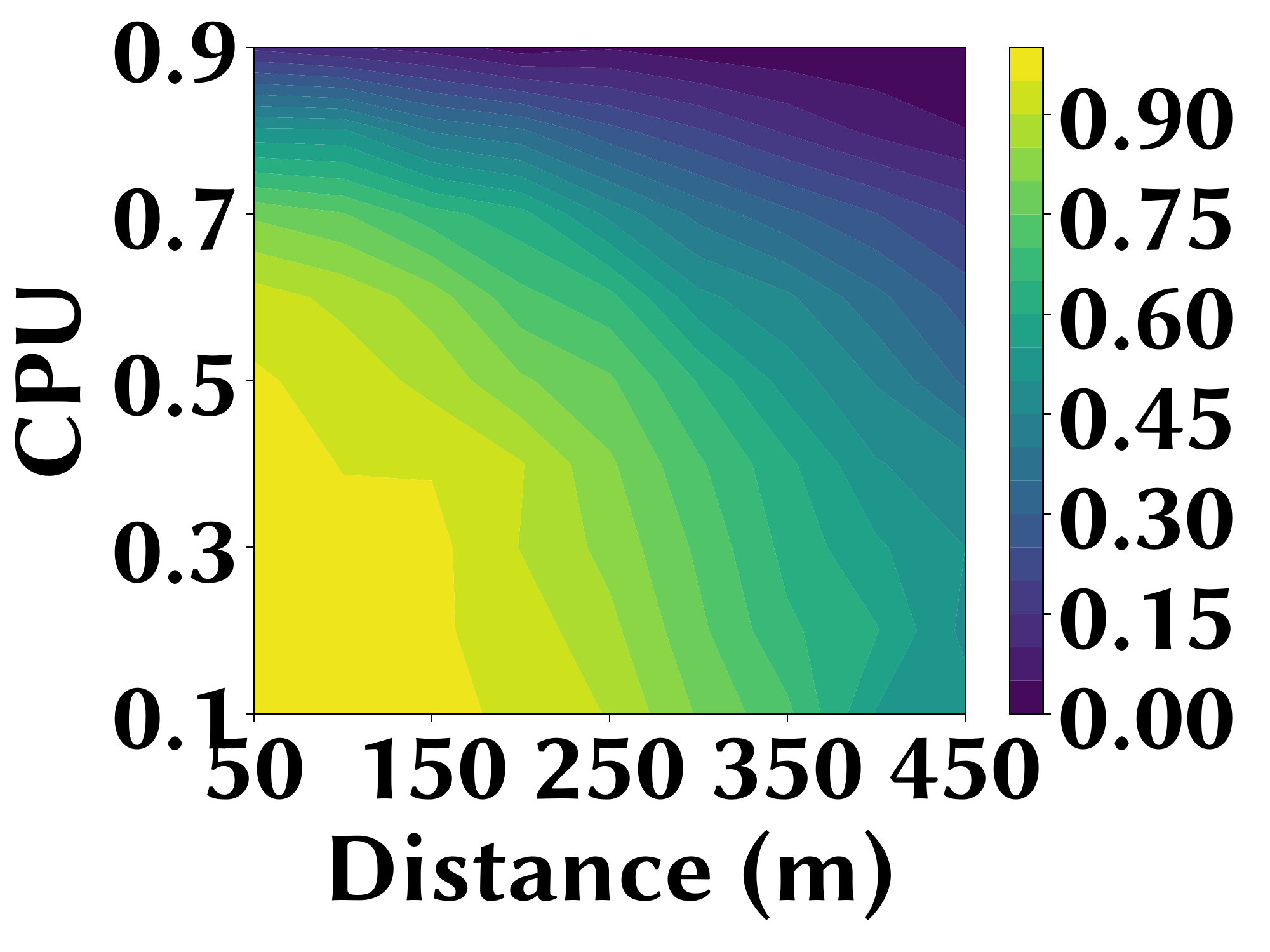}\label{fig:dataset_lt_3}}
	\subfigure[$L=1.0$]{\includegraphics[trim=0cm 0.5cm 0cm 0cm,clip,  width=0.23\textwidth]{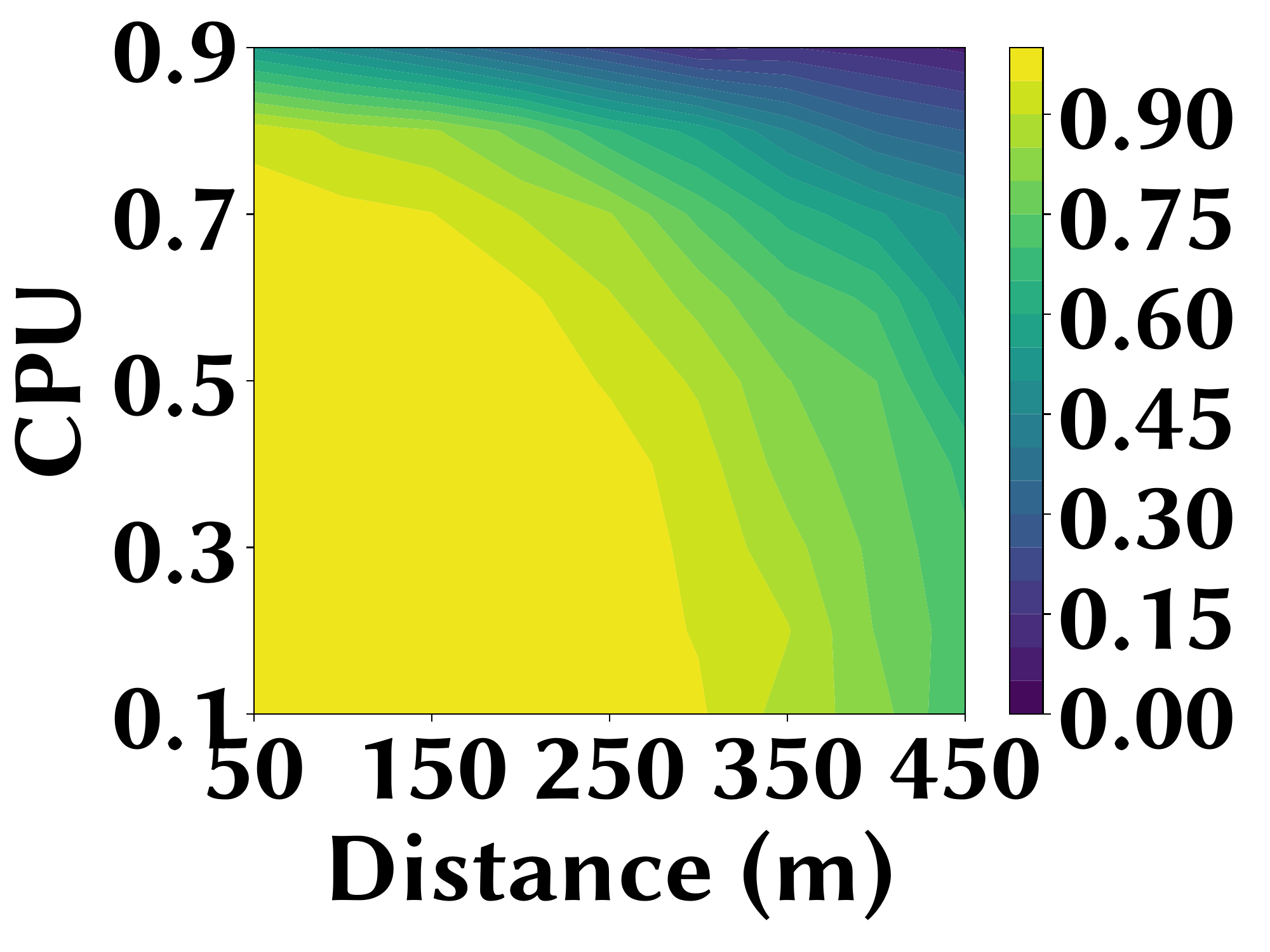}\label{fig:dataset_lt_4}}
	\vspace{-0.3cm}
	\caption{Simulation result for the ground-truth success rate (i.e., context parameter in \ouralg) under different features.}\label{fig:dataset_model_illustrate}
	\vspace{-0.3cm}
\end{figure*}

Besides the success rate, there is also a cost of each offloading (e.g.,
the computational cost incurred, incentives or payment made to the vehicular cloud). Here, we denote $\eta_{ij}$ as the cost for offloading micro service $j$ to vehicular cloud $i$.
Thus, the expected utility/reward given the offloading decision can be formulated as
\begin{equation}\label{eqn:vcc_utility}
U(\mathbf{x},\mathbf{a}) = \prod_{j=1}^M\left[1-\prod_{i=1}^{C}(1-x_{ij} a_{ij})\right]-\sum_{j=1}^{M}\sum_{i=1}^{C} \eta_{ij} a_{ij}
\end{equation}
where $x_{ij}$ is the context parameter, and $a_{ij}$
is the offloading decision, for $i \in \{1,2 ... C\}$ and  $j \in \{1,2 ... M\}$.

In Eqn.~\eqref{eqn:vcc_utility}, the success rate (i.e., context parameter) $x_{ij}$ for each replication is usually predicted by an estimation model based on the environment conditions. We denote the feature as $c_{ij}$, which includes the distance $l$ from RSU to the vehicular cloud, the CPU utilization $cpu$ of the vehicular cloud, and task's deadline.
Naturally, no matter how accurately $x_{ij}$ is estimated based on the feature $c_{ij}$,
there exist estimation errors. Simply ignoring the uncertainty in $x_{ij}$ can result
in arbitrarily bad consequences. Thus, to provide a level of assurance,
we need to consider robust optimization formulated as follows:
\begin{equation}\label{eqn:opt_vec_c4}
\max_{\mathbf{a}\in\mathcal{A}} \min_{\delta\in\Delta} \;U(\mathbf{x}+\delta,\mathbf{a}),
\end{equation}
where we consider an $L_2$-norm uncertainty set  $\Delta=\{\delta,\; |\delta|_2\leq\epsilon\}$. While the greedy algorithm has a performance
guarantee in terms of competitive ratio
due to the submodularity of the objective function
 when the context parameter
$\mathbf{x}$ is perfect  \cite{Combinatorial_book_Korte_2012},
our robust combinatorial optimization problem in Eqn.~\eqref{eqn:opt_vec_c4}
is much more challenging and the greedy algorithm can have a poor performance
(see Section~\ref{sec:large_error}).
Finally, when using \ouralg, we add a minus sign
in the utility function in Eqn.~\eqref{eqn:opt_vec_c4}
to convert the maximin problem into the standard minimax
form.

\subsection{Simulation Setup}

We now describe how we  estimate the context parameter,
calculate the \emph{ground-truth} context in our simulation, and prepare the dataset.

\textbf{Context parameter estimation.}
We estimate the context parameter based on the respective feature information.
In a real-world application, the selection of prediction models is determined by the trade-off between interpretability and accuracy.
For example, a simple linear model is explainable but often has a higher prediction error, whereas a neural network model can predict with a lower error, but lacks good interpretability. In our study, we consider
two prediction models: empirical linear model and residual model implemented with a neural network. More details about prediction models are presented in the appendix.
The linear  model captures the high uncertainty budget case
(i.e., larger $\epsilon$), whereas the neural
network residual model corresponds to a low uncertainty budget.

\textbf{Latency model.} A micro service offloaded to $v_i$ is successful when the computation finishes within the deadline $L$, i.e., $d=d^t+d^c\leq L$. The transmission time $d^t$ can be calculated based on the  wireless channel model as follows:
\begin{equation}\label{eqn:wireless_model_c4}
d^t = \frac{S_{data}}{W\cdot \log_2\left(1+\frac{P\cdot(d)^{-\alpha}}{\sigma^2+I^t}\right)}
\end{equation}
where $S_{data}$ is the offloading data size, $P$ is the transmission power, $\alpha$ is the channel gain factor with respect to the RSU-vehicle distance, $\sigma^2$ is the random noise, and $I^t$ indicates the interference noise.
We set the parameter as followings: $S_{data}=3Mb$, $W=10M$,$P=10dBm$,$\sigma^2=-172dBm$, $I^t\in[-10,-30]dBm$ and $\alpha=1.8$ in our simulation. Besides, the  measurement error for the vehicle's location is set as 3\% (10m) according to the recent GPS manual \cite{gps_5th_edition_2020_c4}.
Then, the CPU-related computation time is simulated and obtained based on the M/M/1 queueing process \cite{M_M_1_queue_model_1974_introduction} using the  dataset in \cite{cpu_latency_dataset_2017_c4}. Concretely, our computation latency model follows $d^c=\frac{a}{b-cpu}+noise\simeq\frac{0.227}{2.15-cpu}+0.007\cdot\mathcal{N}(0,1)$,
where $cpu$ is the CPU utilization.

\textbf{Dataset preparation.} By utilizing the latency model described above, we can generate the {ground-truth} context parameter dataset with the resulting success rate. The success rate for each offloading decision is calculated as the average over $1000$ simulation rounds. The results are illustrated in Fig.~\ref{fig:dataset_model_illustrate} with different features. In total, by considering random
features (i.e., distance, CPU, and deadline), we generate a training dataset with 15k instances, a validation dataset with 4k instances, and a test dataset with 6k instances.

Each task includes $M=4$ micro services, each offloaded to up to  $C=5$ vehicle clouds,
resulting a complexity of $\mathcal{O}(2^{20})$. Thus, each problem instance in our dataset contains a 20-dimensional context vector. Based on the context prediction models
in the appendix, we investigate two error budgets for robustness --- 0.71 for the empirical model, and 0.27 for the neural network model.

\subsection{Baseline Algorithms, Performance Metrics, and Training}
\label{sec:metrics}

\textbf{Baselines.} We consider the following three baseline algorithms and two oracles for comparison.

$\bullet$ \random: The offloading decisions are randomly made without considering the context
parameter or uncertainties.

 $\bullet$ \greedyP: It greedily solves the problem, starting from zero
 decisions (i.e. $a_{ij}=0$) and based on the predicted context.
 
 $\bullet$ \basealgo (Learning for Combinatorial Optimization): It is similar to \ouralg, but oblivious of context uncertainty and directly uses the predicted utility (i.e., $U(\mathbf{x},\mathbf{a})$) to train a neural network for optimization.

$\bullet$ \weak: It uses exhaustive search to maximize the predicted utility $U(\mathbf{x},\mathbf{a})$ without considering uncertainty.

$\bullet$ \oracle: It knows the true context and uses exhaustive search to directly maximize the true utility.

Note that due to  non-convexity of
the inner optimization problem of Eqn.~\eqref{eqn:opt_vec_c4},
constructing another oracle that
exhaustively
searches for the optimal robust decision (e.g.,
using a SOTA solver to solve the inner problem for each sampled candidate decision)
is beyond our computational capability.

\textbf{Performance Metrics.} We consider the following metrics.

$\bullet$ Predicted utility: It is the utility
by viewing the predicted context $\mathbf{x}$ as the true one.

$\bullet$ True utility: It is the utility under the true context.

$\bullet$ Worst-case utility: It is the worst-case utility over the entire
the context uncertainty set.

The worst-case context may not be the true one. Thus, in general, there exists
a tradeoff between worst-case utility and the true utility --- more emphasis on
the worst-case robustness performance can compromise the average performance.
On the other hand,
the predicted utility
has no practical meaning.
In this paper, we focus on robustness, while also showing the predicted
and true utility for reference.

During the testing phase, to avoid biases of using our own maximizer when calculating the worst-case utility (i.e., our maximizer may favor \ouralg),
we use the solver in \textit{scipy.optimize} package to replace
our maximizer and solve the inner optimization problem. The decision in \ouralg
is still made by using our own maximizer.
In other words, given a decision $\mathbf{a}$ (by any baseline/oracle algorithm, or \ouralg),
we use the solver to calculate $\min_{\delta\in\Delta} \;U(\mathbf{x}+\delta,\mathbf{a})$.
Note that the solver is much slower (i.e., $\sim$70x in our experiment)
than running forward propagation in our ensemble of maximizer networks.

\begin{figure}[!t]
	\centering
	\includegraphics[trim=0cm 8.2cm 11.7cm 0cm,clip,  width=0.4\textwidth]{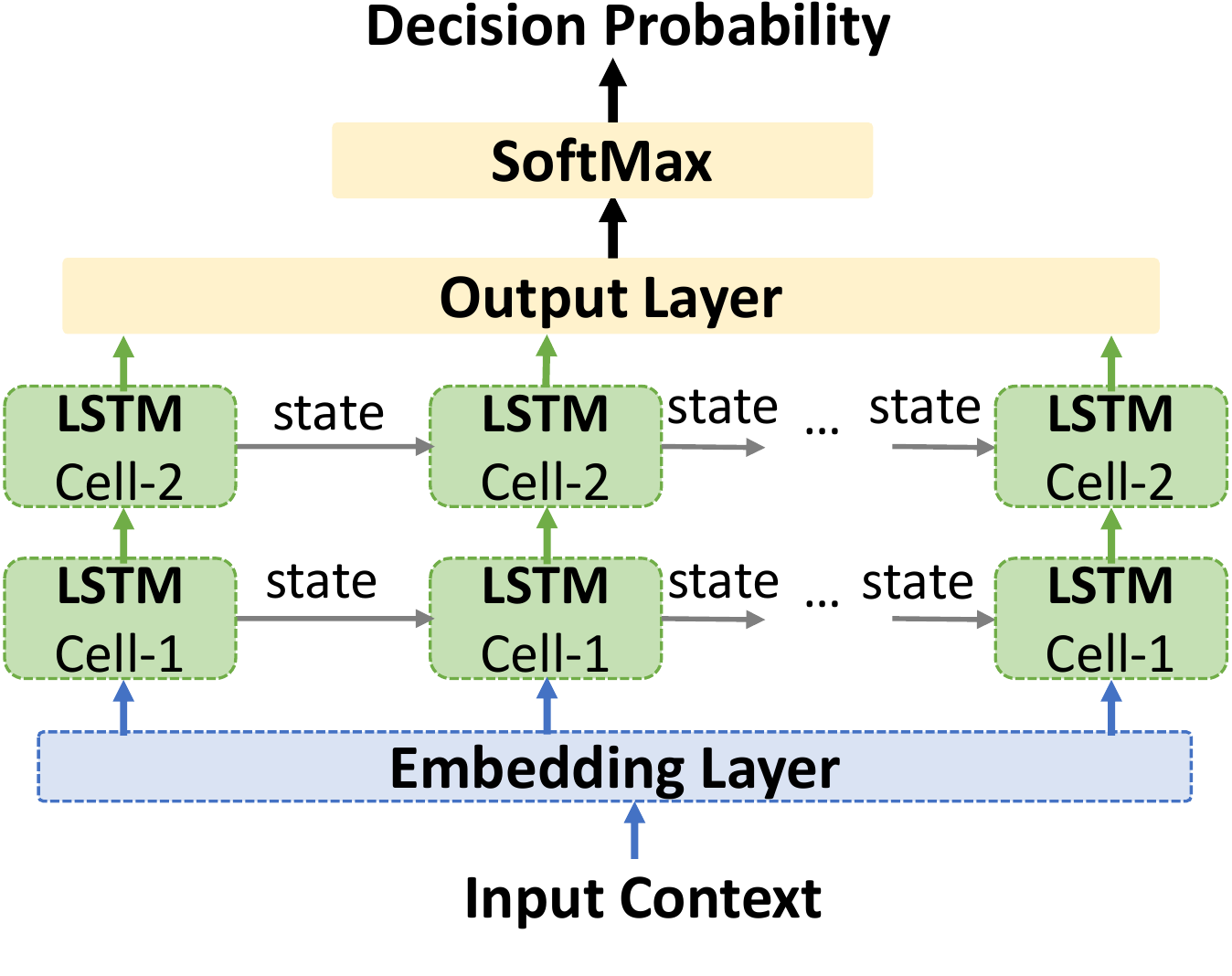}
	\vspace{-0.4cm}
	\caption{{Architecture of the
minimizer network in our simulation.}}\label{fig:mini_net}
\end{figure}

\textbf{Training.} {The minimizer network is implemented with
a neural network with an embedding layer, two LSTM layers with {cell-1 with 50 hidden nodes and cell-2 with one hidden node}, and a fully-connected output layer with $N=20$ nodes and softmax activation. The detailed architecture is illustrated in Fig.~\ref{fig:mini_net}. The model is trained by Adam optimizer \cite{adam_optimizer_2014}. To avoid gradient explosion, an exponentially decreasing learning rate and gradient capping are employed. More specifically, the learning rate is initialized as $1e^{-3}$ and decreases by a factor of 0.9 for every 20 epochs. In the inference/tesing phase, by default, {1000 decisions} are sampled based on the output probability $P_{\theta_a}(\mathbf{a}|\mathbf{x})$, and an optimal decision is selected by the maximizer.}
Note that  \basealgo is also implemented with the same network architecture, but trained by directly using the predicted utility $U(\mathbf{x},\mathbf{a})$.

\begin{table}[!t]
	\caption{Average utility 
		under a {large} error budget.}
	\label{table:lin_performance_result}
	\setlength{\tabcolsep}{9pt}
	\vspace{-0.15cm}
	\centering
	\begin{tabular}{|l|c|c|c|}
		\hline
		\textbf{Algorithm} & \textbf{Predicted} & \textbf{True} & \textbf{Worst Case} \\
		\hline
		\randombf & 0.3408 & 0.1398 & -0.0694 \\
		\greedybfP   & 0.7395 & 0.4103 & 0.1075 \\
		\basealgobf & 0.7545 & 0.4289 & 0.1206 \\
		\weakalgobfP& \textbf{0.7812} & 0.4441 & 0.1234 \\
		\oraclealgobfP & 0.7055 & \textbf{0.6345} & 0.1600 \\
		\hline
		\ouralgobfP & {0.6481} & {0.5357} & {\textbf{0.3940}}\\
		\hline
	\end{tabular}
\end{table}

We train an ensemble of four neural networks to solve the inner worst-case part in \eqref{eqn:opt_vec_c4}.
According to Eqn.~\eqref{eqn:opt_vec_c4}, the performance is only impacts by $\delta_{i,j}$ and $\mathbf{a}_{i,j}$.
Thus, each network includes  two hidden layers, each with 400 neurons, and one {customized output} layer, whose node value is multiplied by $\mathbf{a}_{i,j}$
(such that $\delta_{ij}=0$ when $\mathbf{a}_{i,j}=0)$.
These networks are trained using the same dataset, but different
initial weights and weight parameter $\lambda$ in the loss function in Eqn.~\eqref{eqn:inner_max_loss}.

\subsection{Results with A Large Error Budget}\label{sec:large_error}
We now investigate the performance of \ouralg with a large context parameter error budget $\Delta=\{\delta, |\delta|_2\leq0.71\}$, when an empirical linear model is used for
success rate prediction.

We show the average performances of different algorithms on the test dataset summarized in Table~\ref{table:lin_performance_result}. The worst-case utility is our focus, while
the true utility reflects how well an algorithm performs on average in the typical cases.
The results show that  \ouralgoP provides the best performance among all algorithms (including the oracle methods) under the worst-case condition, demonstrating its strong robustness. Meanwhile, the true utility of \ouralgoP outperforms the other algorithms, except for \oraclealgoP that knows the true context in advance. \weak has the highest
predicted utility because it directly maximizes the predicted utility, but the predicted
utility has no practical meaning.
Importantly, by comparing \ouralg with \basealgo,
we see that being oblivious of the context uncertainty can take a high toll
in terms of the worst-case utility and robustness.
Moreover,
 \basealgo can provide near-optimal predicted utility compared with \weakalgoP, confirming
  that the learning-based optimizer can be used to solve combinatorial optimization.

\begin{figure}[!t]
	\centering
	\subfigure[]{\includegraphics[trim=0cm 0cm 0cm 0cm,clip,  width=0.24\textwidth]{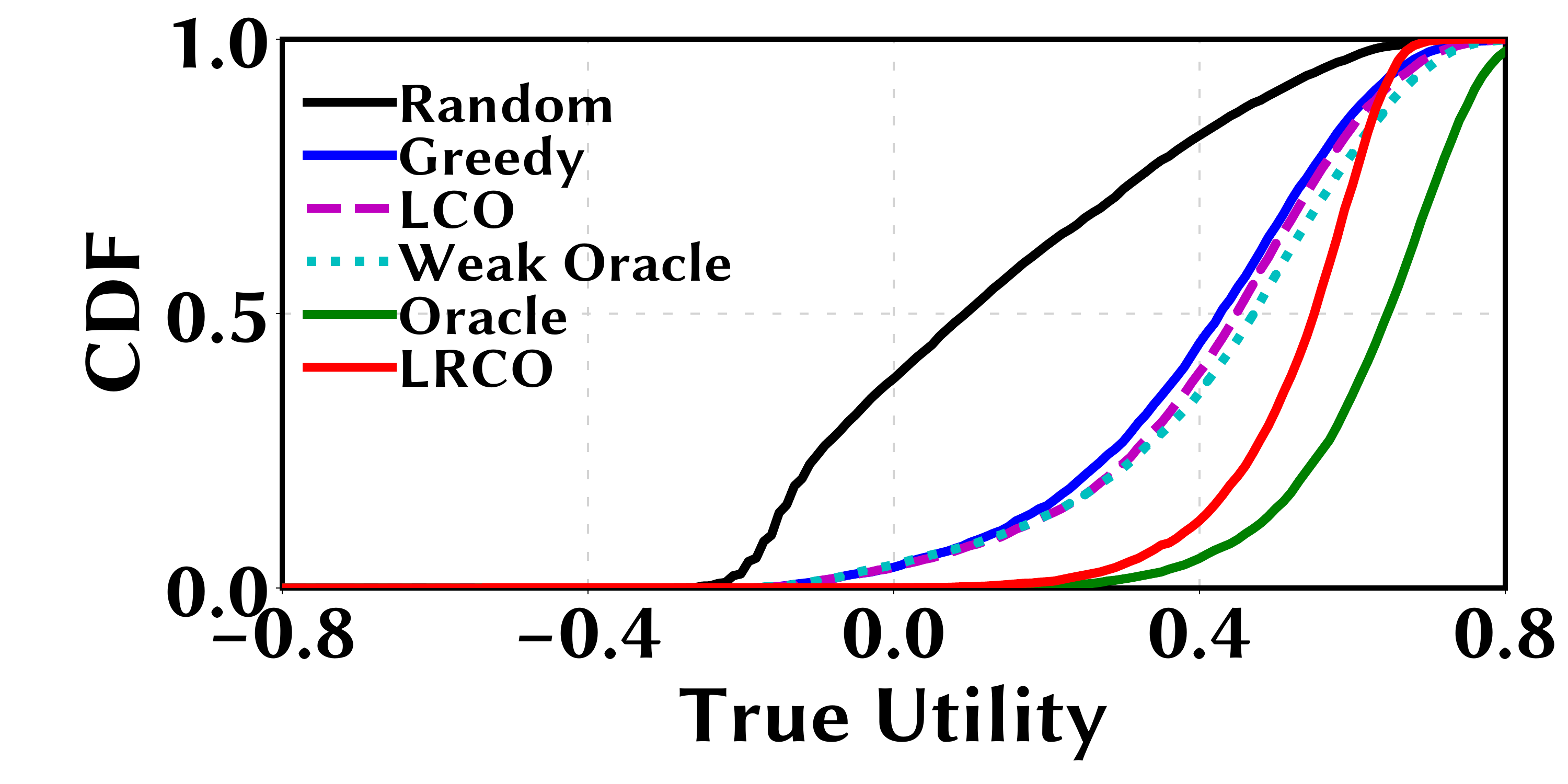}\label{fig:linear_result_y0_cdf}}
	\subfigure[]{\includegraphics[trim=0cm 0cm 0cm 0cm,clip,  width=0.24\textwidth]{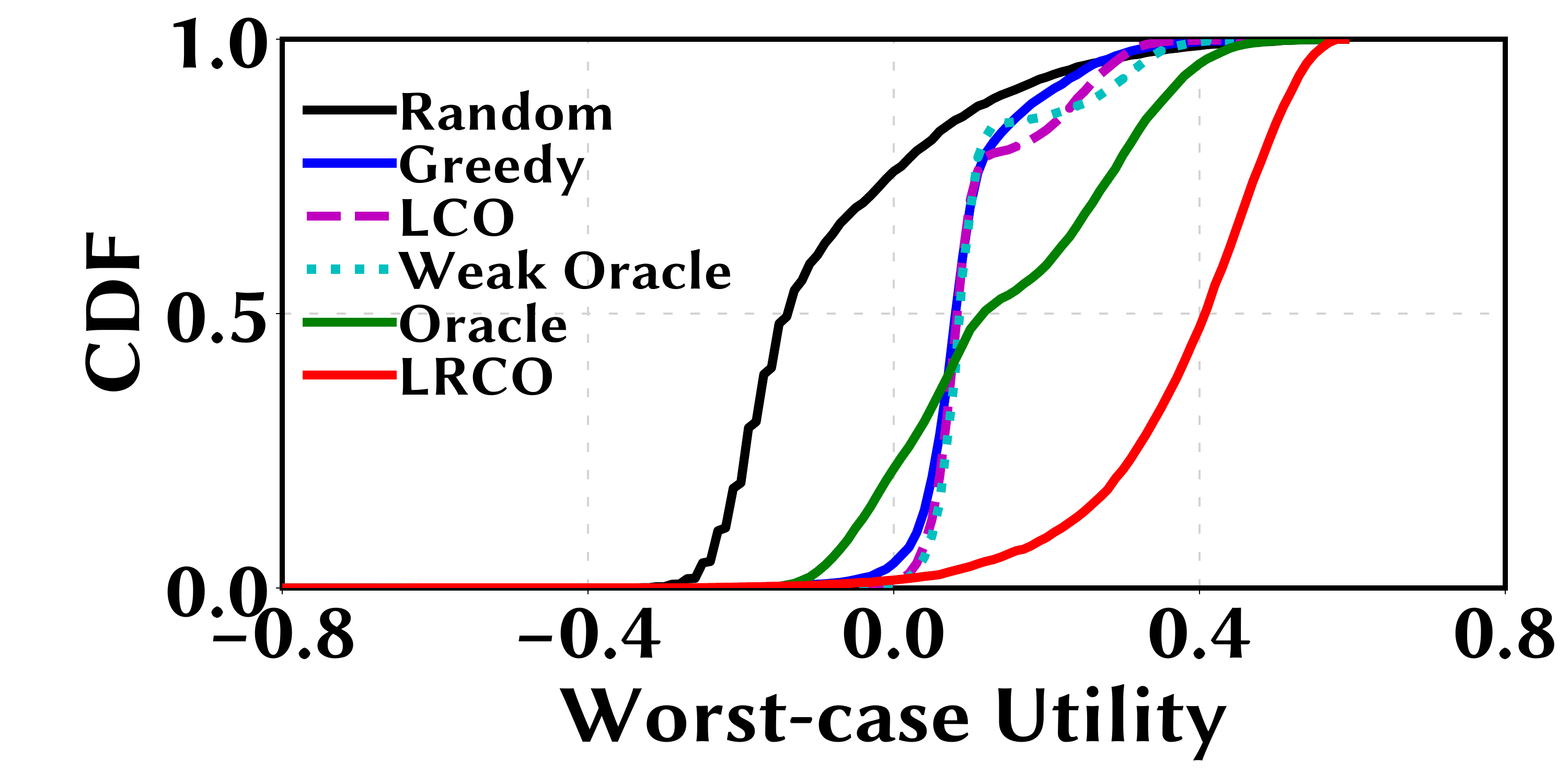}\label{fig:linear_result_sotay_cdf}}
	\vspace{-0.5cm}
	\caption{CDF plots of utilities under a {large} error budget. (a) True utility. (b) The worst-case utility.}\label{fig:lin_performance_cdf}
	\vspace{-0.3cm}
\end{figure}

Additionally, we visualize in Fig.~\ref{fig:lin_performance_cdf} the cumulative distribution function (CDF)
of true utilities and  worst-case utilities for different algorithms.
These results show that \ouralg can provide near-optimal true utility
performance, although there is a gap due to the consideration of robustness
in the presence of a large error budget.
Further,  \ouralgoP can provide better robust performance than oracle algorithms.

\begin{table}[!t]
	\caption{Average utility under a {small} error budget.}
	\label{table:res_performance_result}
	\setlength{\tabcolsep}{9pt}
	\vspace{-0.15cm}
	\centering
	\begin{tabular}{|l|c|c|c|}
		\hline
		\textbf{Algorithm} & \textbf{Predicted} & \textbf{True} & \textbf{Worst Case}  \\
		\hline
		\randombf & 0.1564 & 0.1398 & -0.0022 \\
		\greedybfP   & 0.6150 & 0.5760 & 0.3763 \\
		\basealgobf & 0.6421 & 0.5988 & 0.3988 \\
		
\weakalgobfP & \textbf{0.6553} & 0.6117 & 0.4165 \\
		\oraclealgobfP & 0.6331 & \textbf{0.6345} & 0.4176 \\
		\hline
		\ouralgobfP & {0.5988} & {0.5810} & {\textbf{0.4597}} \\
		\hline
	\end{tabular}
\end{table}

\subsection{Results with A Small Error Budget}\label{sec:small_error}
Now, we turn to the case of a small error budget $\Delta=\{\delta, |\delta|_2\leq0.27\}$, by
using a neural network context predictor described in Appendix.
All other settings remain unchanged.

\begin{figure}[!t]
	\centering
	\subfigure[]{\includegraphics[trim=0cm 0cm 0cm 0cm,clip,  width=0.24\textwidth]{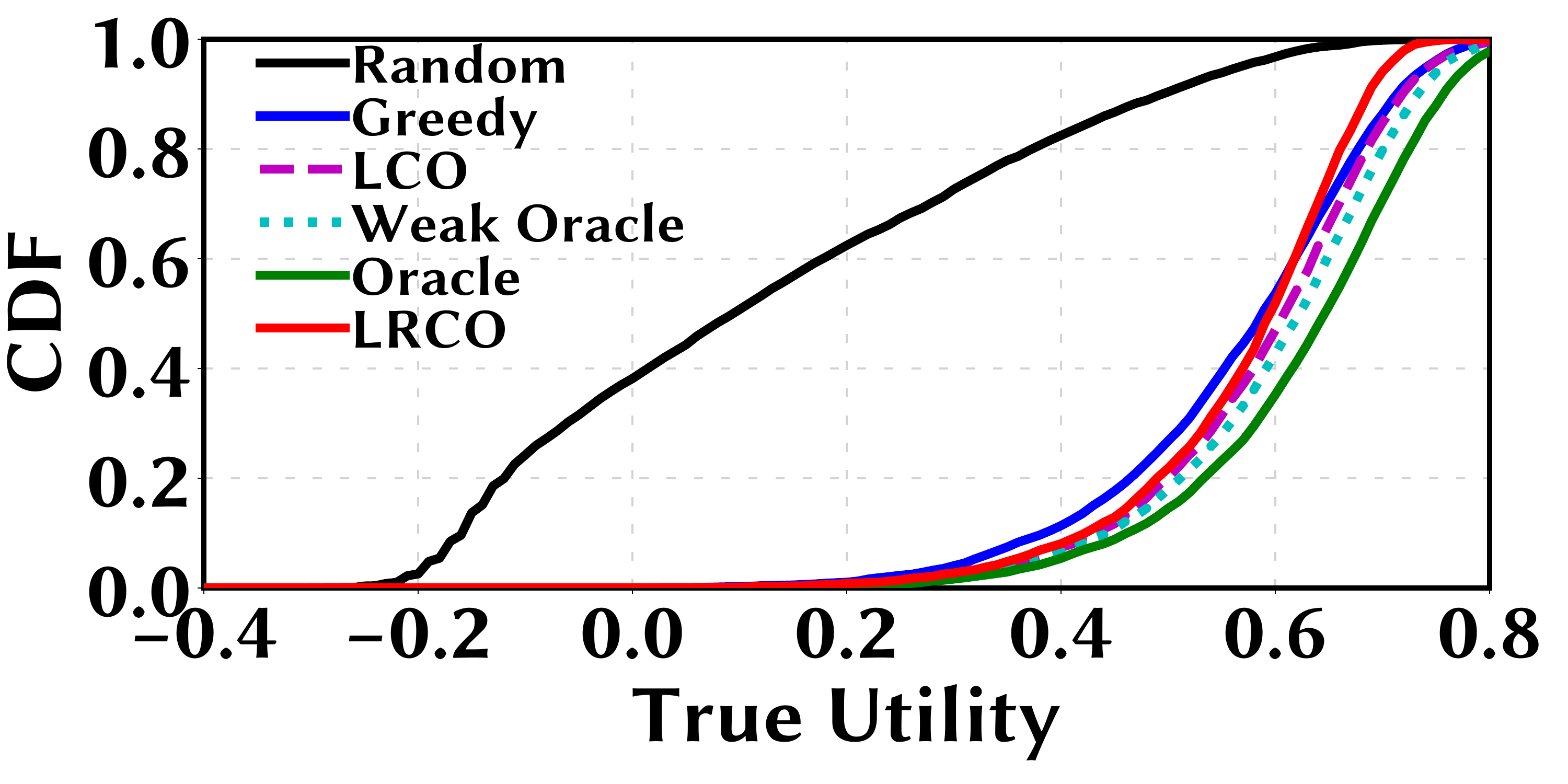}\label{fig:res_result_y0_cdf}}
	\subfigure[]{\includegraphics[trim=0cm 0cm 0cm 0cm,clip,  width=0.24\textwidth]{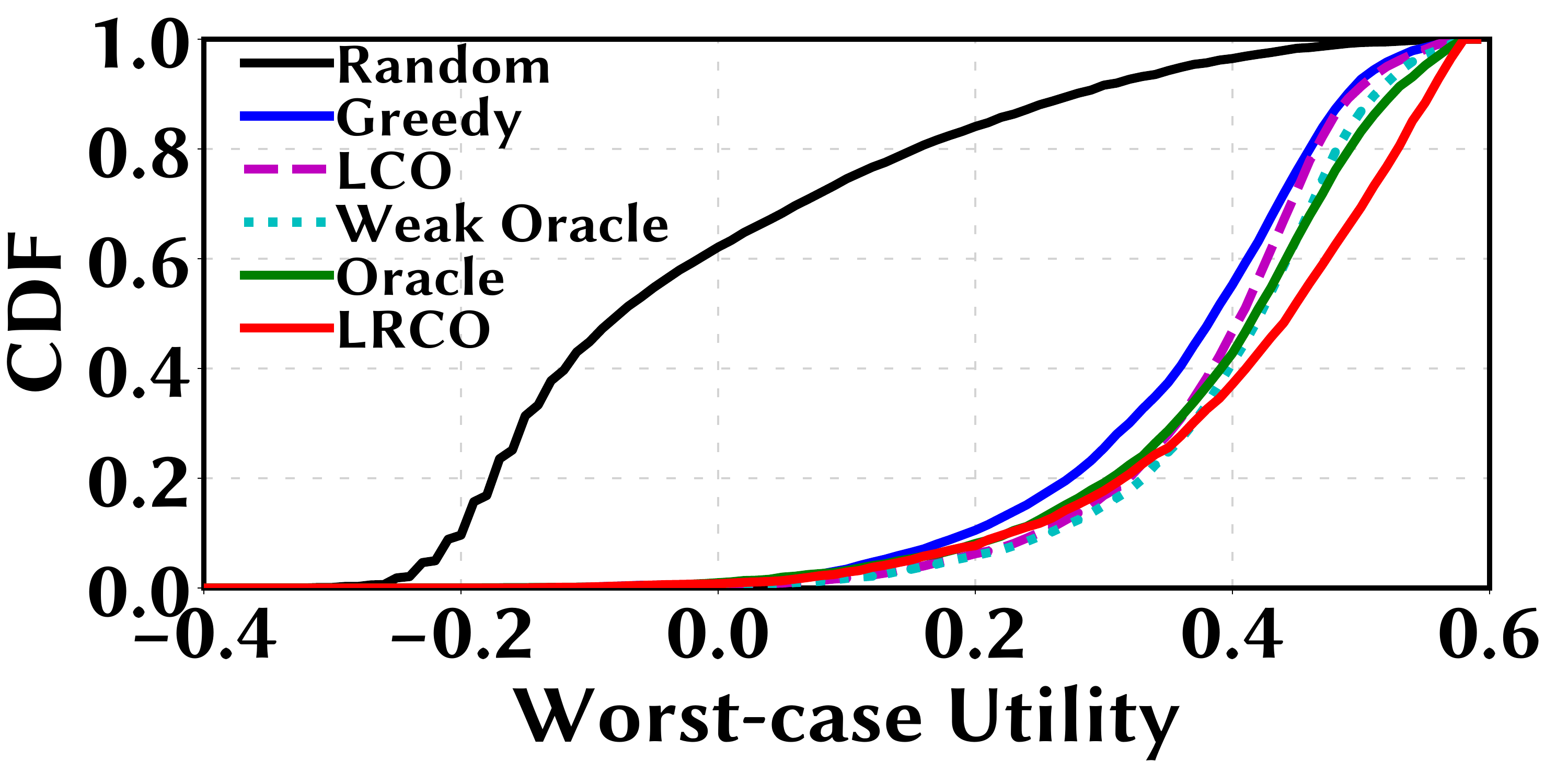}\label{fig:res_result_sotay_cdf}}
	\vspace{-0.3cm}
	\caption{CDF plots of utilities under a {small} error budget. (a) True utility. (b) The worst-case utility.}\label{fig:res_performance_cdf}
	\vspace{-0.3cm}
\end{figure}

We show in Table~\ref{table:res_performance_result}  the average utility under different algorithms and metrics. Also, the utility CDF is shown in Fig.~\ref{fig:res_performance_cdf}. In the case
of a small context prediction error, we have similar observations
as in the case of a large prediction error in Section~\ref{sec:large_error}.
As intuitively expected, the  gap between the true utility and the worst-case utility becomes smaller, since the context uncertainty set is smaller.
Importantly, among all the algorithms, \ouralg still has the highest worst-case utility and robustness.

\subsection{Sensitivity Study}\label{sec:sensitivity}
We perform the sensitivity study for
$\Delta=\{\delta, |\delta|_2\leq0.71\}$, while the results for $\Delta=\{\delta, |\delta|_2\leq0.27\}$ are also similar and hence
omitted due to space limitation.

\textbf{Impact of ensemble settings.} In \ouralgoP, the maximizer is comprised of an ensemble of four neural networks, considering the non-convex nature of $U(\mathbf{x},
\mathbf{a})$. 
Here, we present the comparion
results with and without  ensembling in Table~\ref{table:impact_ensemble}, including two $\lambda$ settings. We see that the ensemble approach provides a better
and more robust performance, although having a single neural
network for the maximizer only has minor degradation.

\begin{table}[!t]
	\caption{Ensemble settings for \ouralgoP.}
	\label{table:impact_ensemble}
	\vspace{-0.1cm}
	\setlength{\tabcolsep}{9pt}
	\centering
	\begin{tabular}{|l|c|c|c|}
		\hline
		\textbf{Setting} & \textbf{Predicted} & \textbf{True} & \textbf{Worst Case} \\
		\hline
		Ensemble (default) & 0.6481 & 0.5357 & 0.3940 \\
		\hline
		w/o Ensemble ($\lambda=1$)   & 0.6423 & 0.5289 & 0.3842 $\downarrow$ \\
		\hline
		w/o Ensemble ($\lambda=10$)  & 0.6396 & 0.5274 & 0.3830 $\downarrow$ \\
		\hline
	\end{tabular}
	\vspace{-0.3cm}
\end{table}

\textbf{Impact of minimizer network size.} We  vary the hidden nodes to 20 and 200. The results are shown in Table~\ref{table:impact_model_size}, demonstrating that a larger
network can provide a slightly better performance although the training and inference
time can also increase.

\begin{table}[!t]
	\caption{Minimizer network size for \ouralgoP.}
	\label{table:impact_model_size}
	\vspace{-0.1cm}
	\setlength{\tabcolsep}{9pt}
	\centering
	\begin{tabular}{|l|c|c|c|}
		\hline
		\textbf{\# of Hidden Nodes} & \textbf{Predicted} & \textbf{True} & \textbf{Worst Case} \\
		\hline
		20 & 0.6436 & 0.5299 & 0.3888 $\downarrow$ \\\hline
		50 (default)& 0.6481 & 0.5357 & 0.3940 \\\hline
		200 & 0.6480 & 0.5358 & 0.3941 \\
		\hline
	\end{tabular}
	\vspace{-0.3cm}
\end{table}

\begin{figure}[!t]
	\centering
	\subfigure[]{\includegraphics[trim=0cm 0cm 0cm 0cm,clip,  width=0.24\textwidth]{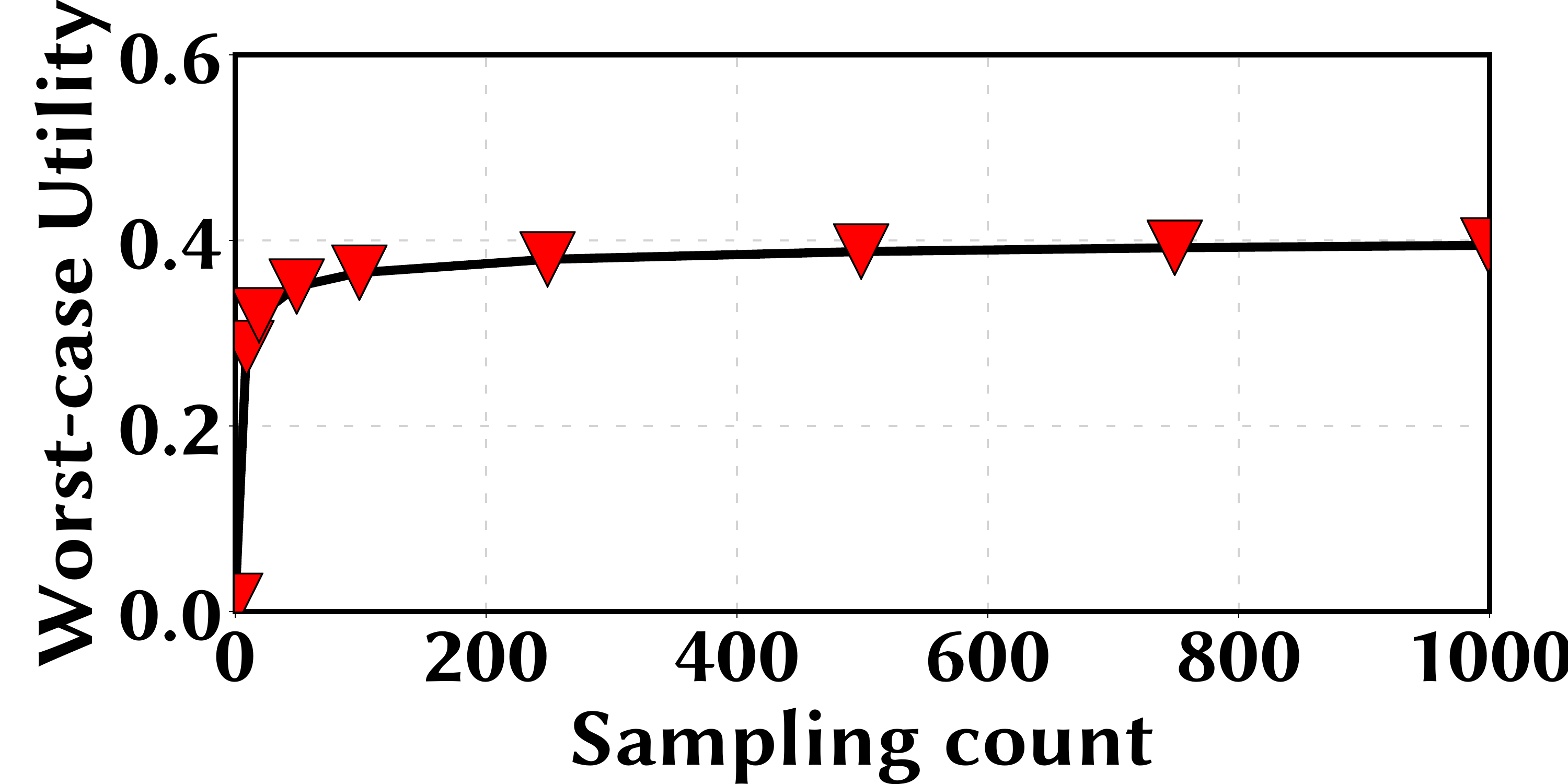}\label{fig:sampling_sensitivity}}
	\subfigure[]{\includegraphics[trim=0cm 0cm 0cm 0cm,clip,  width=0.24\textwidth]{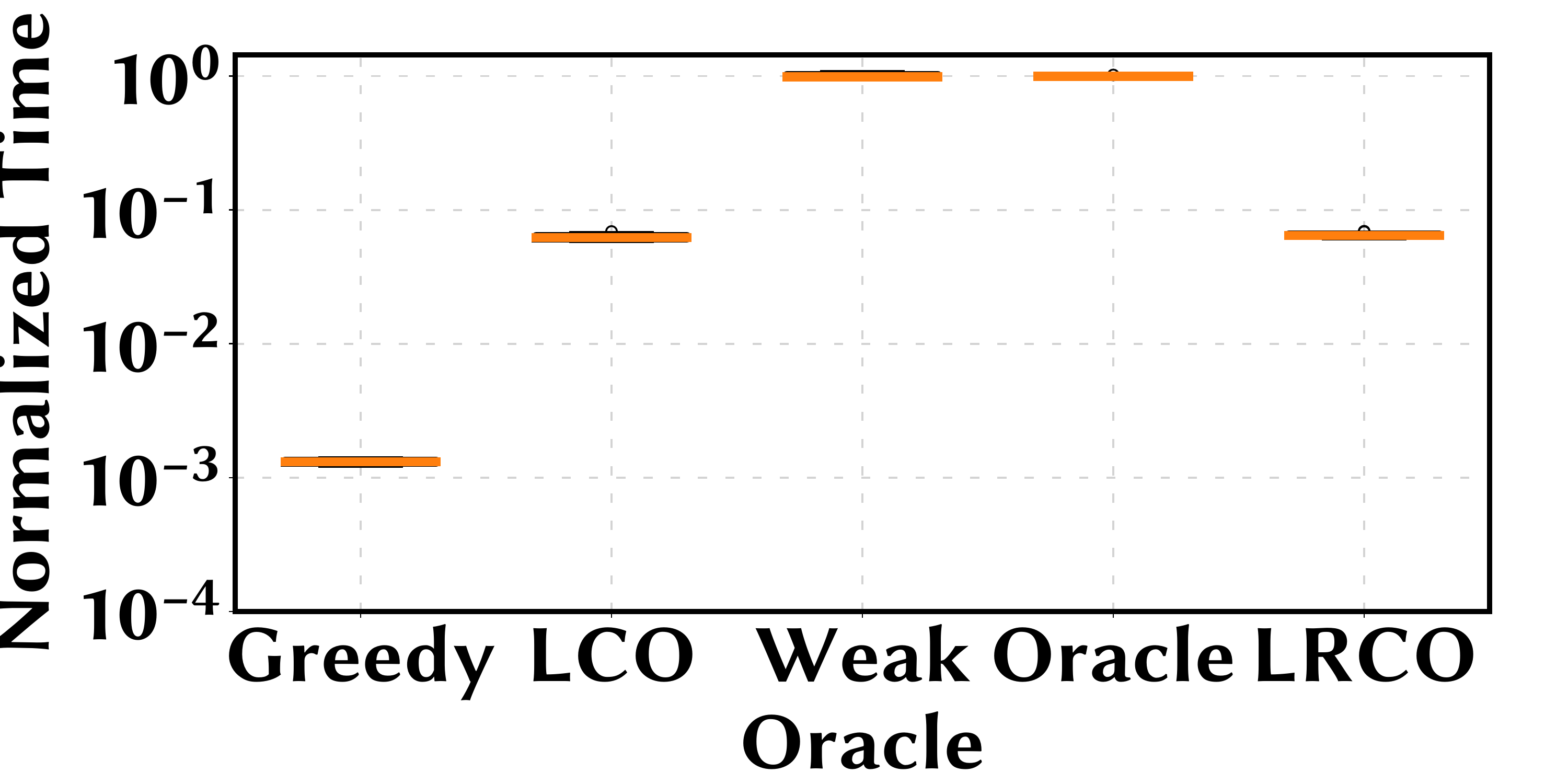}\label{fig:time_analysis}}
	\vspace{-0.3cm}
	\caption{(a) Performance with different sampling counts. (b) Inference time complexity  for different algorithms.}\label{fig:sensitivity}
	\vspace{-0.3cm}
\end{figure}

\textbf{Impact of sampling count.} In the testing phase, by default, the best decision is selected by sampling 1000 candidate decisions based on the probability produced by the minimizer network. Now, we show the performance with different numbers of samples. The robust performance is shown in Fig.~\ref{fig:sampling_sensitivity}. In general,
the more samples, the better performance. This comes at a cost of increasing
the inference time, as more candidate decisions need to be evaluated for the worst-case utility
by our maximizer in \ouralg.

\subsection{Inference Time}
A key advantage of \ouralg compared with the existing solvers is the fast inference time
at runtime, although it incurs extra training cost offline.
The time is recorded with \textit{`time'} package in Python3. We run 10 times over all the testing data for each algorithm. Finally, the average recorded time over 10 runs is normalized with respect to the time of \oraclealgoP.
The normalized time is illustrated with box-plot in Fig.~\ref{fig:time_analysis}. The results demonstrate that \ouralg can provide compelling solutions much faster ($\sim${10x}) than oracle algorithms. While \greedyP is fast and can provide a good performance for the predicted utility, it is not robust.
Also, we see that \ouralg is only slightly slower than \basealgo, showing
that the forward propagation through our neural networks in the maximizer
is very fast. Note further that the solver in \textit{scipy.optimize} package is much slower (i.e., $\sim$70x) than our neural network. Thus, it is beyond our computational capability
to use the solver along with an exhaustive search method -- for each testing problem instance, approximately $2^{20}$ runs of the solver
are needed to find the optimal robust decision in our experiment.
\section{Conclusion}\label{sec:c4_conclusion}
In this work, we have studied robust combinatorial optimization
and proposed \ouralg, a learning-based optimizer that quickly outputs a robust solution in the presence of uncertain context.
\ouralg leverages a pair of learning-based optimizers --- one for the minimizer and
the other for the maximizer.
 Finally, to evaluate the performance of \ouralg,
we have performed simulations for
the  task offloading problem in vehicular edge computing.
Our results highlighted that \ouralg can greatly improve robustness, while having a very low runtime complexity.

\section*{Acknowledgement}
Z. Shaolei, J. Yang, and S. Ren are supported in part by the NSF under
grants CNS-1551661 and CNS-1910208. C. Shen is supported in
part by the NSF under grant SWIFT-2029978.

\bibliographystyle{ieeetr}

\appendix[Context Parameter Prediction]

We consider two context parameter prediction
models, which predict the success rate $x$ based on the feature: $x=M_{p}(d,cpu,L)$. The first model is implemented with linear regression: ${x}=w_dd+w_{cpu}cpu+w_LL+bias$, which is easy to interpret. Additionally, we provide a neural network-based residual model, including the pre-trained linear model and a parallel residual block. The residual block is implemented with a neural network to learn the residual error.

In the training phase, both linear and neural network models are trained with the same training dataset.
Specifically, the neural network residual block includes two hidden layers, each with 20 neurons, and \textit{`ReLu'} activation. The residual block is trained by minimizing the \textit{`mean squared error'} loss function using the Adam optimizer (learning rate $1e-4$).

\begin{figure}[!h]
	\centering
	\subfigure[Estimation error]{\includegraphics[trim=0cm 0cm 0cm 0cm,clip,  width=0.24\textwidth]{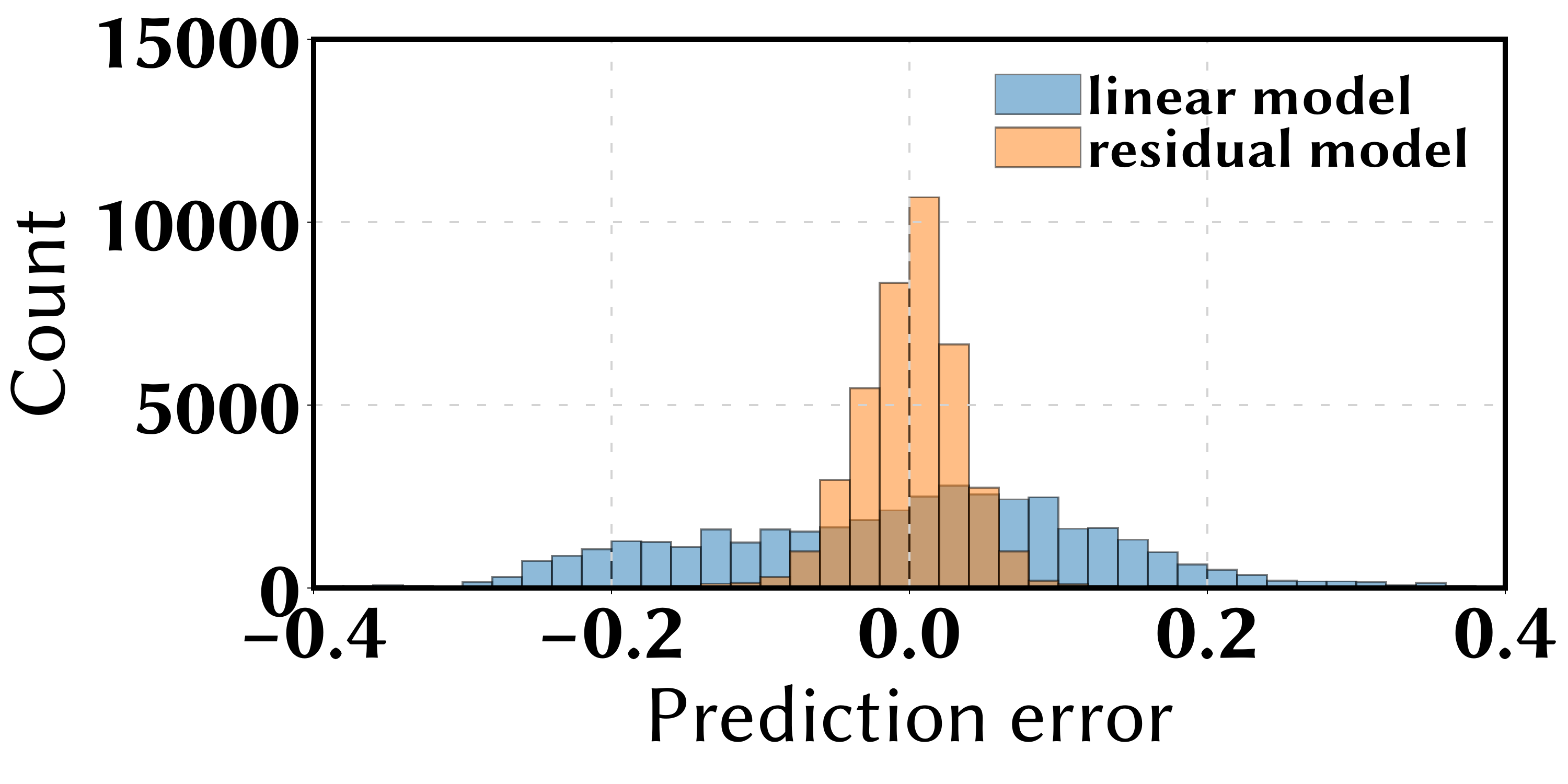}\label{fig:x_individual_error}}
	\subfigure[L2 norm of estimation error]{\includegraphics[trim=0cm 0cm 0cm 0cm,clip,  width=0.24\textwidth]{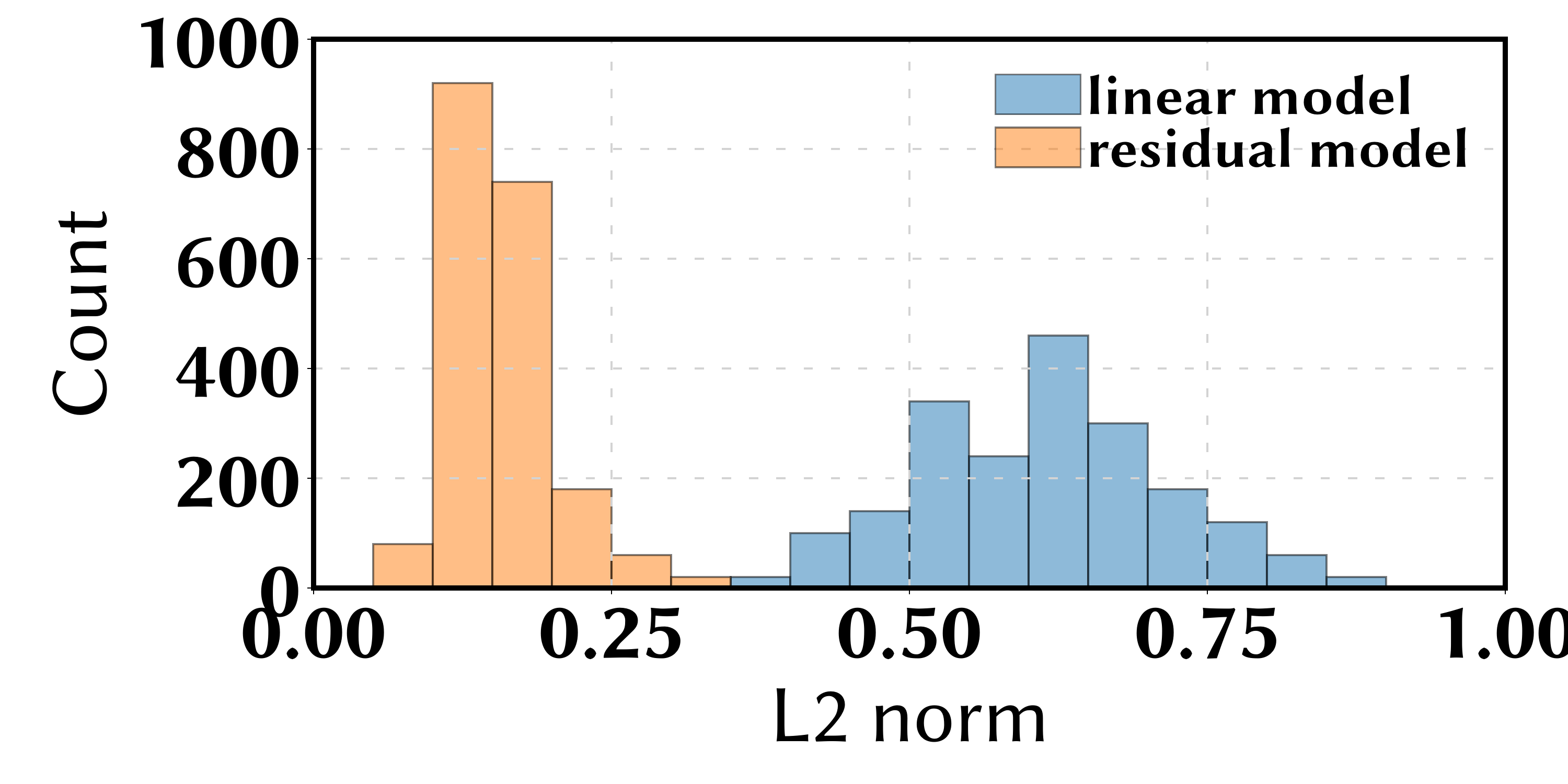}\label{fig:x_l2_budget}}
	\vspace{-0.5cm}
	\caption{Context parameter (success rate) estimation error and its $L_2$ norm.}\label{fig:x_estimation_error}
	\vspace{-0.2cm}
\end{figure}
The predicted success rate by the linear model is denoted as ${x}^l$, while ${x}^r$ represents the result of the residual model. Then, two errors are investigated: the element-wise prediction error  and the $L_2$ norm of 20 elements in the context parameter. We show the results in Fig.~\ref{fig:x_estimation_error}. The results validate that the residual model indeed reduces the estimation error. Thus, we will consider the robust optimization problem with both a large error budget and a small error budget. Here, we consider the 99-percentile $L_2$ norm as the error budget for robustness, which is 0.71 for the empirical model and 0.27 for the residual model, respectively.

\end{document}